\documentclass[10pt, a4paper, logo]{googledeepmind} %
\usepackage{times}
\usepackage{makecell}
\usepackage{amssymb}

\usepackage{hyperref}
\usepackage[capitalise,noabbrev]{cleveref}
\usepackage[leftcaption]{sidecap} 

\usepackage{amsmath,amsfonts,bm}
\usepackage{natbib}
\usepackage{fancyhdr}

\def\eqref#1{equation~\ref{#1}}

\def\1{\bm{1}}

\DeclareMathAlphabet{\mathsfit}{\encodingdefault}{\sfdefault}{m}{sl}
\SetMathAlphabet{\mathsfit}{bold}{\encodingdefault}{\sfdefault}{bx}{n}

\usepackage{url}
\usepackage{booktabs}
\usepackage{graphicx}
\usepackage{subcaption}
\usepackage{amssymb}
\usepackage[most]{tcolorbox}
\usepackage[table]{xcolor}
\tcbuselibrary{breakable} 
\usepackage{amsmath}
\usepackage{siunitx}
\usepackage{colortbl}

\usepackage{microtype}
\usepackage{color}
\usepackage{wrapfig}
\usepackage{natbib}
\usepackage{colortbl}
\usepackage{microtype}
\usepackage{graphicx}
\usepackage{booktabs} 
\usepackage{array}
\usepackage{textcomp}
\usepackage{stfloats}
\usepackage{float}
\usepackage{verbatim}
\usepackage[ruled, linesnumbered, lined]{algorithm2e}
\usepackage{multirow}
\usepackage{enumitem}

\definecolor{BestColor}{HTML}{C8E6C9}  
\definecolor{SecondBestColor}{HTML}{FFF9C4} 

\usepackage{pifont}
\DeclareUnicodeCharacter{2713}{\ding{51}} 

\newcommand{\greencheck}{\textcolor{ggg}{\textbf{✓}}} 
\newcommand{\redx}{\textcolor{rrr}{\textbf{X}}}       

\newcommand{\best}[1]{\cellcolor{BestColor}\textbf{#1}}
\newcommand{\secondbest}[1]{\cellcolor{SecondBestColor}#1}
\usepackage{tcolorbox}
\usepackage{amsmath,amsfonts}

\definecolor{ggg}{RGB}{26,179,0}
\definecolor{rrr}{RGB}{179,0,0}
\definecolor{oodc}{RGB}{31,73,121}
\definecolor{idc}{RGB}{68,142,68}

\definecolor{mygray}{gray}{0.9}

\def\Bias#1#2{\bm{b}}

\newtcolorbox{examplebox}[2][]{ 
    breakable, 
    enhanced, 
    colback=white, 
    colframe=cyan, 
    coltitle=white, 
    fonttitle=\bfseries, 
    title=#2, 
    overlay middle={\draw[cyan, line width=1pt](frame.south west)--(frame.south east);}, 
    overlay last={\draw[cyan, line width=1pt](frame.south west)--(frame.south east);}, 
    #1 
}

\usepackage{amssymb} 
\usepackage[T1]{fontenc}
\usepackage{booktabs}      
\usepackage{graphicx}      
\usepackage[table]{xcolor} 
\usepackage{siunitx}       
\usepackage{etoolbox}      
\usepackage[normalem]{ulem}     

\definecolor{impcolor}{HTML}{2E8B57} 

\newcommand{\improvementstyle}[1]{$^{\textcolor{impcolor}{\tiny #1}}$}

\newcommand{\scoreimp}[2]{%
  \textbf{#1}%
  \ifstrequal{#2}{+0.0}{}{%
    \ifstrequal{#2}{0.0}{}{%
      \makebox[0pt][l]{\improvementstyle{#2}}%
    }%
  }%
}

\title{Rationale Matters: Learning Transferable Rubrics via Proxy-Guided Critique for VLM Reward Models}

\author[1,]{Weijie Qiu\textsuperscript{$\spadesuit$}\textsuperscript{*}}
\author[1]{Dai Guan\textsuperscript{$\spadesuit$}}
\author[2]{Junxin Wang}
\author[1]{Zhihang Li$^\dagger$}
\author[1]{Yongbo Gai}
\author[1]{Mengyu Zhou}
\author[1]{Erchao Zhao}
\author[1]{Xiaoxi Jiang}
\author[1]{Guanjun Jiang}
\affil[1]{Qwen Large Model Application Team, Alibaba}
\affil[2]{Institute of Automation, Chinese Academy of Sciences}

\begin{abstract}
Generative reward models (GRMs) for vision-language models (VLMs) often evaluate outputs via a three-stage pipeline: rubric generation, criterion-based scoring, and a final verdict. However, the intermediate rubric is rarely optimized directly. Prior work typically either treats rubrics as incidental or relies on expensive LLM-as-judge checks that provide no differentiable signal and limited training-time guidance.

We propose \textbf{Proxy-GRM}, which introduces \textbf{proxy-guided rubric verification} into Reinforcement Learning (RL) to explicitly enhance rubric quality. Concretely, we train lightweight proxy agents (\textbf{Proxy-SFT} and \textbf{Proxy-RL}) that take a candidate rubric together with the original query and preference pair, and then predict the preference ordering using only the rubric as evidence. The proxy's prediction accuracy serves as a \textbf{rubric-quality reward}, incentivizing the model to produce rubrics that are internally consistent and transferable.
With \({\sim}50\text{k}\) data samples, Proxy-GRM reaches state-of-the-art results on the VL-Reward Bench, Multimodal Reward Bench, and MM-RLHF-Reward Bench, outperforming the methods trained on four times the data. Ablations show Proxy-SFT is a stronger verifier than Proxy-RL, and implicit reward aggregation performs best. Crucially, the learned rubrics transfer to unseen evaluators, improving reward accuracy at test time without additional training. Our code is available at \url{https://github.com/Qwen-Applications/Proxy-GRM}.
\end{abstract}

\begin{document}
\maketitle
\section{Introduction}
\label{sec:introduction}
\begin{figure*}[t]
  \centering
  \includegraphics[width=\textwidth]{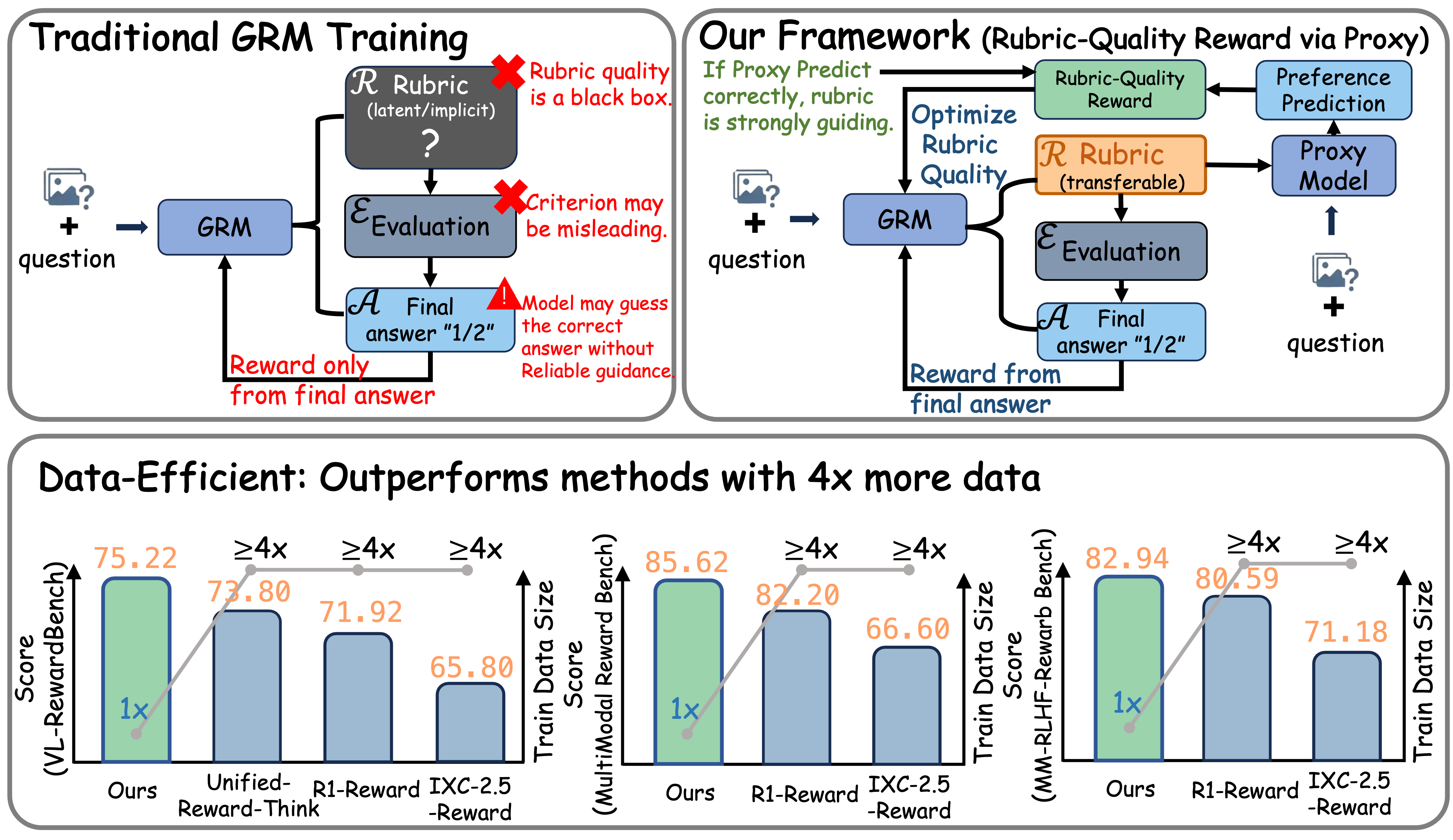}
  \caption{\textbf{Overview of rubric-quality reward modeling via a proxy.} \textbf{Top-left:} Traditional GRM training uses reward only from the final answer; rubric quality is implicit/latent, making the supervision a black box and allowing the model to guess correct answers without reliable guidance. \textbf{Top-right:} Our framework explicitly generates a transferable rubric alongside the final answer and optimizes the GRM with a rubric-quality reward computed by a proxy model that predicts preference given the question and generated rubric; correct proxy predictions provide strong guidance to improve rubric quality. \textbf{Bottom:} Data-efficiency comparison on VL-RewardBench, MultiModal Reward Bench and MM-RLHF-Reward Bench shows our method (1\(\times\) data) outperforming baselines trained with \(\geq\)4\(\times\) more data.}
  \label{fig:compare}
\end{figure*}

Vision-language models (VLMs)~\citep{alayrac2022flamingo} have advanced rapidly across a broad range of multimodal tasks, including visual question answering, image captioning, and complex multimodal reasoning~\citep{liu2023visual,bai2023qwen}. As these models are increasingly deployed in open-ended generation settings, reliably evaluating and guiding the quality of their outputs has become a central challenge~\citep{ouyang2022training,zheng2023judging,lambert2025rewardbench}. Early approaches adopted \textbf{scalar reward models} that compress response quality into a single numeric score; while efficient, such models offer little interpretability---they reveal \emph{whether} a response is preferred but not \emph{why}~\citep{ouyang2022training}. To overcome this, \textbf{Generative Reward Models} (GRMs) produce natural-language critiques alongside a preference verdict~\citep{zhang2024generative,ankner2024critique,xiong2025llava} as shown in \Cref{fig:compare}, typically structured as:
\begin{equation}
\texttt{<rubric>}\mathcal{R}\texttt{</rubric><eval>}\mathcal{E}\texttt{</eval><answer>}\mathcal{A}\texttt{</answer>}
\end{equation}
\noindent where \(\mathcal{R}\) is a set of evaluation criteria (the \emph{rubric}), \(\mathcal{E}\) is a criterion-by-criterion assessment, and \(\mathcal{A}\in\{1,2\}\) is the final preference verdict~\citep{kim2023prometheus,kim2024prometheus}. This decomposition endows GRMs with both structure and interpretability.

Despite the popularity of this paradigm, the reliability of the entire pipeline depends critically on the quality of its first step: the rubric \(\mathcal{R}\). A rubric that is vague, incomplete, or subtly biased renders the subsequent critique meaningless. The problem is compounded by how GRMs are trained: existing methods~\citep{wang2025unifiedthink,shen2025vlm} optimize exclusively for the \emph{final outcome}---assigning a reward to \(\mathcal{A}\) or computing a loss on the verdict token---leaving the intermediate rubric entirely \textbf{unsupervised and unverified}. Consequently, models may learn to produce rubrics that \emph{rationalize} a pre-determined answer rather than \emph{guide} principled evaluation, analogous to post-hoc rationalization in chain-of-thought reasoning~\citep{turpin2023language,lanham2023measuring}. Such rubrics exhibit poor \textbf{generalizability and transferability}: when applied by an independent evaluator or on a different task distribution, they fail to elicit accurate preference judgments.

The most common remedy is \textbf{LLM-as-a-Judge}: prompting an external model to score the generated rubric~\citep{zheng2023judging,youssef2024llms,chen2023adaptation}. This suffers from two fundamental limitations. First, it is computationally expensive and impractical at the training scale. Second, and more critically, it \textbf{cannot close the training loop}: an external judgment is not differentiable and cannot serve as a reinforcement learning reward for the GRM itself, leaving rubric quality perpetually open-loop. Recent text-only efforts have explored learning rubrics from annotations~\citep{xie2025auto} or using rubrics as structured reward signals~\citep{gunjal2025rubrics,xu2026alternating}, but none address closed-loop rubric verification in the multimodal setting.

We ask: \textbf{Can we learn to generate rubrics verified by an independent agent, with this signal integrated into the training loop?} We answer affirmatively with \textbf{Proxy-GRM}, which introduces a dedicated \emph{proxy evaluator} measuring rubric \emph{transferability}---the degree to which a rubric, when given to an independent model, continues to elicit correct preference judgments. We train two proxy variants, \textbf{Proxy-SFT} and \textbf{Proxy-RL}, using their agreement signal as a closed-loop reward. A key finding is that \textbf{Proxy-SFT surprisingly outperforms Proxy-RL}, revealing that outcome-only RL supervision can degrade process-level evaluation fidelity. Proxy-GRM achieves state-of-the-art performance on VL-RewardBench, Multimodal Reward Bench, and MM-RLHF-Reward Bench using only \({\sim}50\text{k}\) training samples---\(4\times\) less data than comparable methods---and its rubrics transfer effectively to unseen evaluator models without additional training.

Our main contributions are as follows:
\begin{itemize}[leftmargin=*]
    \item We identify \textbf{rubric transferability} as a critical and under-explored dimension of multimodal GRM quality, and propose \textbf{proxy-based verification} as a principled, scalable, and training-loop-compatible solution.
    \item We propose \textbf{Proxy-GRM}, a closed-loop training framework that integrates an independent proxy evaluator as a rubric quality signal for RL, directly addressing the open-loop limitation of LLM-as-a-Judge approaches.
    \item We train two proxy variants (\textbf{Proxy-SFT} and \textbf{Proxy-RL}) and show that Proxy-SFT outperforms Proxy-RL, revealing a tension between outcome-level RL optimization and process-level evaluation fidelity.
    \item We achieve \textbf{state-of-the-art} results on VL-RewardBench, Multimodal Reward Bench, and MM-RLHF-Reward Bench with only \({\sim}50\text{k}\) training samples (\({\approx}4\times\) less than comparable methods), with rubrics that generalize to unseen evaluators without retraining.
\end{itemize}

\section{Related Work}
\label{sec:related_work}

\subsection{Reward Models for Vision-Language Models}
Reward models serve as a critical component in aligning large language models and vision-language models with human preferences~\citep{christiano2017deep,stiennon2020learning,bai2022training}. Early approaches primarily employed discriminative reward models that output scalar scores~\citep{bradley1952rank,zhu2024starling}. Bradley--Terry models~\citep{ouyang2022training} and their variants have been widely used to parameterize pairwise preferences. In the multimodal domain, models such as IXC-2.5-Reward~\citep{zang2025internlm}, MM-RLHF-Reward~\citep{zhang2025mm}, and LLaVA-Critic~\citep{xiong2025llava} have extended these ideas to handle image-text inputs.

More recently, generative reward models (GRMs)\citep{jia2025writingzerobridgegapnonverifiable,jia2026openrubricsystemscaling} have gained traction due to their ability to produce interpretable critiques alongside preference judgments~\citep{zhang2024generative,ankner2024critique,mahan2024generative}. R1-Reward~\citep{zhang2025r1} introduced stable reinforcement learning techniques for training multimodal reward models with chain-of-thought reasoning. Unified-Reward~\citep{wang2025unified} proposed a unified framework that handles both text and multimodal evaluation through structured thinking. These models typically follow the rubric-evaluation-answer decomposition described in Section~1, but none of them explicitly verify or optimize the quality of the generated rubric.

\subsection{Rubric-Based Evaluation and Reward Modeling}
The use of rubrics structured evaluation criteria has a long history in educational assessment~\citep{brookhart2013create} and has recently been adopted in LLM evaluation. JudgeLM~\citep{zhu2023judgelm} and Prometheus~\citep{kim2023prometheus,kim2024prometheus} pioneered the use of fine-grained rubrics for LLM-as-judge evaluation. Auto-Rubric~\citep{xie2025auto} proposed learning implicit rubric weights from human preference data and converting them to explicit, interpretable rubrics for reward modeling. This work highlighted the importance of rubric quality but focused on extracting rubrics from existing annotations rather than generating and verifying them end-to-end.

In the reinforcement learning context, Rubrics-as-Rewards~\citep{gunjal2025rubrics} demonstrated that rubric-structured rewards can extend RL beyond verifiable domains (e.g., math) to subjective tasks like creative writing and open-ended dialogue. Their approach uses rubrics to decompose complex evaluation into manageable sub-criteria, each scored independently. Alternating RL for Rubric-based Reward Modeling~\citep{xu2026alternating} further advanced this direction by proposing an alternating optimization scheme between rubric refinement and reward model training for non-verifiable LLM post-training.

CritiqueVLM~\citep{wu2025alignmmbench} explored critique-based evaluation for multimodal models but did not address rubric transferability. RubricEval~\citep{bhat2023rubriceval} proposed using rubrics for systematic evaluation of language models, but focused on benchmark design rather than rubric generation quality. Our work differs fundamentally from all prior art in that we (i) introduce an external proxy agent to verify rubric transferability during training, and (ii) integrate this verification signal into the RL loop as a reward component, enabling end-to-end optimization of rubric quality.

\subsection{Reinforcement Learning for Language and Vision-Language Models}
Reinforcement learning from human feedback (RLHF)~\citep{christiano2017deep,ziegler2019fine} and its variants---including Direct Preference Optimization (DPO)~\citep{rafailov2023direct}, Group Relative Policy Optimization (GRPO)~\citep{shao2024deepseekmath}, and Proximal Policy Optimization (PPO)~\citep{schulman2017proximal}---have become standard tools for aligning generative models. In the VLM domain, RLHF-V~\citep{yu2024rlhf} and RLAIF-V~\citep{yu2024rlaif} demonstrated effective preference learning for reducing hallucination and improving response quality.

Recent work has explored using RL to train the reward models themselves. R1-Reward~\citep{zhang2025r1} employed stable RL training with carefully designed reward signals for multimodal reward models. Our approach extends this line of research by introducing a novel proxy-based reward component that evaluates the quality of the model's intermediate reasoning (rubrics) rather than only its final output, bridging the gap between outcome supervision and process supervision~\citep{lightman2023let,uesato2022solvingmathwordproblems}.

\subsection{Process Supervision and Verification}
The distinction between outcome supervision and process supervision has been extensively studied in mathematical reasoning~\citep{lightman2023let,wang2024math}. Process reward models (PRMs) evaluate intermediate reasoning steps, providing denser feedback than outcome reward models (ORMs) that only evaluate final answers. Our proxy-based rubric verification can be viewed as a form of process supervision applied to the rubric generation step: rather than only rewarding correct final verdicts, we additionally reward the generation of rubrics that are independently verifiable. This perspective connects our work to recent advances in self-verification~\citep{weng2023large} and critic models~\citep{mcaleese2024llm,gou2023critic}, while introducing the novel element of external verification through proxy agents.

\begin{figure*}[t]
  \centering
  \includegraphics[width=\textwidth]{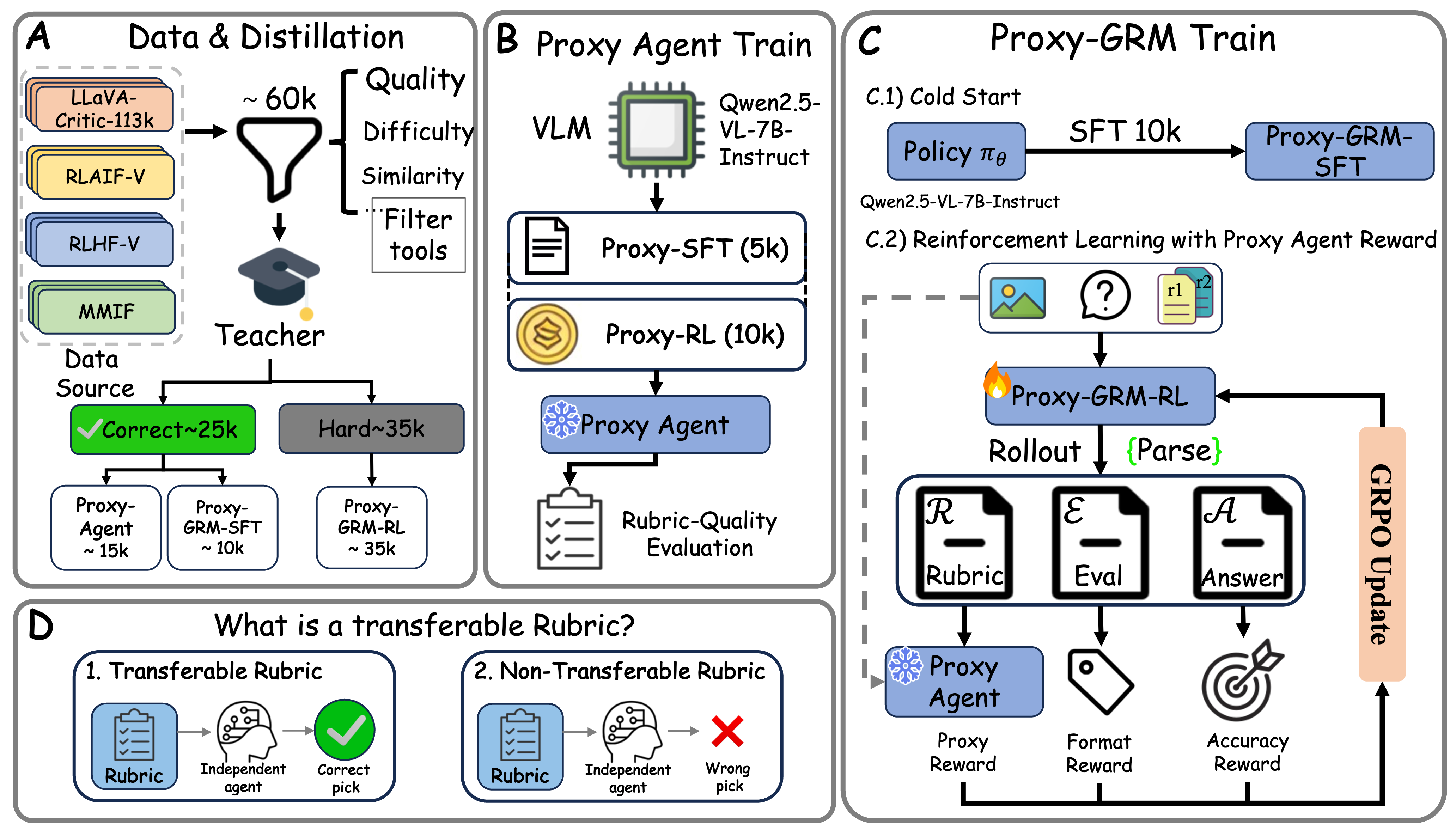}
  \caption{\textbf{Training pipeline for Proxy-GRM with transferable rubrics.} \textbf{(A)} Data \& distillation. We merge multiple VLM preference datasets and apply automated filters to obtain \({\sim}60\text{k}\) samples. A teacher then separates the data into Correct and Hard subsets, which are allocated to train the proxy agent and to cold-start / RL-train Proxy-GRM. \textbf{(B)} Proxy agent training. Starting from Qwen2.5-VL-7B-Instruct, we train a proxy agent via Proxy-SFT followed by Proxy-RL, producing an evaluator that provides rubric-quality assessment signals. \textbf{(C)} Proxy-GRM training. We first cold-start the GRM policy with SFT. We then perform reinforcement learning with accuracy, proxy (rubric-quality), and format rewards, and optimize with GRPO updates. \textbf{(D)} Transferable rubric. A rubric is transferable if an independent agent can follow it to select the correct answer; otherwise, it is non-transferable and may lead to wrong decisions despite appearing plausible.}
  \label{fig:framework}
\end{figure*}

\section{Methods}
\label{sec:methods}
\vspace{-1mm} 

We present Proxy-GRM, a framework for training multimodal generative reward models with transferable rubric generation. Our approach consists of two main components: (1) proxy agent training for rubric evaluation, and (2) proxy-guided reinforcement learning for the policy model. \Cref{fig:framework} provides an overview of the complete pipeline. We first describe the problem formulation (\Cref{sec:problem_formulation}), then detail the proxy agent training (\Cref{sec:proxy_agent_training}), the policy model training (\Cref{sec:policy_model_training}), and the inference procedure (\Cref{sec:inference}).

\subsection{Problem Formulation}
\label{sec:problem_formulation}

Given a multimodal query \(q\) (comprising a textual question and an associated image \(\mathcal{I}\)), along with a pair of candidate responses \((r_1,r_2)\) where one is preferred over the other according to human judgment, a generative reward model \(\pi_\theta\) produces a structured output:
\begin{equation}
    y = \pi_{\theta}(\mathcal{I}, q, r_1, r_2) = 
    \underbrace{\text{\texttt{<rubric>}} \, \mathcal{R} \, \text{\texttt{</rubric>}}}_{\text{evaluation rubric}} 
    \underbrace{\text{\texttt{<eval>}} \, \mathcal{E} \, \text{\texttt{</eval>}}}_{\text{detailed evaluation}} 
    \underbrace{\text{\texttt{<answer>}} \, \mathcal{A} \, \text{\texttt{</answer>}}}_{\text{preference response}}
\end{equation}
where \(\mathcal{R}\) is a set of evaluation criteria (rubric), \(\mathcal{E}\) is the criterion-by-criterion evaluation, and \(\mathcal{A}\in\{1,2\}\) is the index of the preferred response. The model is trained to maximize the probability of producing the correct preference \(\mathcal{A}^*\).

We define \textbf{rubric transferability} as the property that the generated rubric \(\mathcal{R}\) is sufficiently informative and unbiased such that an independent evaluator, given only (\(\mathcal{I}, q, r_1, r_2, \mathcal{R}\)), can also arrive at the correct preference \(\mathcal{A}^*\). Formally, for a proxy agent \(\phi\):
\begin{equation}
    \text{Transferability}(\mathcal{R}) = \bm{1}[\phi(q, \mathcal{I},r_1,r_2,\mathcal{R}) = \mathcal{A}^*]
\end{equation}
This definition captures the intuition that a good rubric should be transferable—it should guide any competent evaluator toward the correct judgment, not just the model that generated it.




\subsection{Proxy Agent Training}
\label{sec:proxy_agent_training}

To operationalize rubric transferability, we introduce a dedicated \textbf{proxy agent} trained exclusively to \emph{consume} rubrics rather than generate them. The proxy receives as input the original query $q$, image $\mathcal{I}$, candidate responses $(r_1, r_2)$, and an externally provided rubric $\mathcal{R}$, and produces a criterion-by-criterion evaluation $\mathcal{E}$ and a final verdict $\mathcal{A} \in \{1, 2\}$ following the <think></think>\textcolor{purple}{\texttt{<answer>}}\textcolor{purple}{\texttt{</answer>}} format without generating a rubric of its own. The proxy's accuracy thus serves as a direct measure of rubric transferability: success indicates that $\mathcal{R}$ encodes sufficient discriminative information; failure signals that $\mathcal{R}$ is underspecified or misleading.

\textbf{Proxy-SFT}. We fine-tune Qwen2.5-VL-7B-Instruct~\citep{bai2025qwen25vltechnicalreport} on $5$k correctly distilled samples via supervised fine-tuning. Each training instance consists of the full input tuple $(q, \mathcal{I}, r_1, r_2, \mathcal{R})$ paired with a ground-truth evaluation and verdict, where the rubric $\mathcal{R}$ is sourced from high-quality annotations. The training objective is standard cross-entropy loss. Proxy-SFT is designed to internalize the skill of faithfully following a given rubric to produce a well-grounded preference judgment, without any incentive to deviate from or reinterpret the provided criteria.

\textbf{Proxy-RL}. Starting from the Proxy-SFT checkpoint, we further train the proxy agent with reinforcement learning on $10$k samples, using binary accuracy—whether the predicted verdict $\mathcal{A}$ matches the ground-truth label $\mathcal{A}^{*}$—as the sole reward signal. This stage aims to improve the proxy's robustness as a rubric consumer under a wider variety of rubric phrasings and quality levels. An important design principle shared by both configurations is that the proxy agents are trained \emph{only} to follow rubrics, not to generate them. This strict specialization ensures that, during policy model training, the proxy serves as a faithful and independent judge of rubric quality, whose verdict is not confounded by its own rubric generation tendencies.

\subsection{Policy Model Training}
\label{sec:policy_model_training}

\Cref{fig:framework} illustrates the overall architecture of Proxy-GRM. Our framework consists of two sequential training stages:

\paragraph{\textbf{Stage 1: Cold-Start Supervised Fine-Tuning.}} We fine-tune a base VLM (Qwen2.5-VL-7B-Instruct) on 10k correctly distilled samples to obtain Proxy-GRM-SFT. This stage teaches the model the structured output format and provides a reasonable initialization for RL:
\begin{equation}
    \mathcal{L}_{\mathrm{cold}}=-\mathbb{E}_{(x,y)\sim\mathcal{D}_{\mathrm{cold}}}\left[\log\pi_\theta(y\mid x)\right]
\end{equation}

\paragraph{\textbf{Stage 2: Proxy-Guided Reinforcement Learning.}} Starting from Proxy-GRM-SFT, we train the policy model using GRPO \citep{shao2024deepseekmath} with a composite reward function that incorporates three components:

\textbf{Accuracy reward} \(r_{\text{acc}}\): evaluates whether the policy model's final verdict is correct. 
\begin{equation}
    r_{\text{acc}} = 
    \begin{cases} 
        +1 & \text{if } \mathcal{A} = \mathcal{A}^* \\ 
        -1 & \text{otherwise }
    \end{cases}
\end{equation}

\textbf{Format reward} \(r_{\text{format}}\): evaluates whether the output follows the prescribed structured format with proper XML tags.
\begin{equation}
    r_{\text{format}} = 
    \begin{cases} 
        +1 & \text{if output matches format}  \\ 
        0 & \text{otherwise } 
    \end{cases}
\end{equation}

\textbf{Proxy reward} \(r_{\text{proxy}}\): evaluates rubric transferability by passing the generated rubric \(\mathcal{R}\) to the frozen proxy agent \(\phi\) and checking whether \(\phi\) can correctly identify the preferred response using only \(\mathcal{R}\):
\begin{equation}
    r_{\text{proxy}} = 
    \begin{cases} 
        +1 & \text{if } \phi(q, \mathcal{I},r_1,r_2,\mathcal{R}) = \mathcal{A}^* \\ 
        -1 & \text{otherwise }
    \end{cases}
\end{equation}
The composite reward is:
\begin{equation}
    r=r_{\text{acc}} + r_{\text{proxy}} + 0.5 \cdot r_{\text{format}}
\end{equation}
This design creates a synergistic training dynamic: \(r_{\text{acc}}\) ensures the policy model learns to make correct judgments, \(r_{\text{proxy}}\) ensures the generated rubrics are robust and transferable, and \(r_{\text{format}}\) maintains structural compliance. The proxy agent is kept frozen throughout policy training, which serves purely as an evaluator and does not co-adapt with the policy.

\subsection{Inference with Proxy Verification}
\label{sec:inference}

At inference time, Proxy-GRM operates in two modes:

\textbf{Standard mode.} The policy model generates a structured output, and the final answer is used directly as the preference verdict. This mode does not require the proxy agent.

\textbf{Proxy-verified mode.} The policy model generates the structured output including rubric \(\mathcal{R}\). The rubric is then passed to the proxy agent along with the original input. If the proxy agent's verdict agrees with the policy model's verdict, the prediction is accepted with high confidence. If they disagree, the policy model's answer is still used (since the proxy agent may also err), but the disagreement can be logged for downstream analysis.

\section{Experiments}
\label{sec:experiments}
\vspace{-1mm} 

We conduct comprehensive experiments to evaluate Proxy-GRM across three dimensions: (1) overall reward modeling performance compared to state-of-the-art methods, (2) analysis studies on proxy agent selection and multi-agent configurations, and (3) rubric transferability to unseen evaluator models.

\subsection{Implementation Details}

\noindent\textbf{Base model.} We use Qwen2.5-VL-7B \citep{bai2025qwen25vltechnicalreport} as the base model for both the policy model (Proxy-GRM) and the proxy agents (Proxy-SFT, Proxy-RL). The teacher model for data distillation is Qwen3-VL-235B-A22B \citep{bai2025qwen3}.

\noindent\textbf{Training data.} We curate ~60k preference samples from LaVA-Critic-113k \citep{xiong2025llava}, RLAIF-V \citep{yu2024rlaif}, RLHF-V \citep{yu2024rlhf}, and MMIF-23k \citep{zhang2025mm}. After distillation and filtering, 25k correct samples are used for proxy agent and policy cold-start training, and the remaining 35k serve as RL training data.

\noindent\textbf{Training details.} For SFT stages of Proxy-SFT and Proxy-GRM, we use a learning rate of \(1 \times 10^{-5}\) with cosine scheduling and train for 1 epoch. For RL stages of Proxy-RL and Proxy-GRM, we use GRPO with a learning rate of \(5 \times 10^{-6}\), group size G=7.

\noindent\textbf{Evaluation benchmarks.} We evaluate on three established multimodal reward benchmarks:
\begin{itemize}[noitemsep, topsep=0pt] 
    \item VL-RewardBench \citep{li2025vl}: A comprehensive benchmark covering general multimodal understanding, hallucination detection, and visual reasoning, with 1,247 preference pairs.
    \item Multimodal Reward Bench \citep{yasunaga2025multimodal}: A diverse benchmark spanning correctness, preference, knowledge, mathematics, coding, safety, and VQA, with 5,000 samples.
    \item MM-RLHF-Reward Bench \citep{zhang2025mm}: A challenging benchmark including multiple-choice, long-form, short-form, safety, and video evaluation tasks.
\end{itemize}

\noindent\textbf{Baselines.} We compare against: (1) closed-source models including GPT-4o, Claude-3.5/3.7-Sonnet, and Gemini-1.5-Pro/Flash; (2) open-source reward models including IXC-2.5-Reward \citep{zang2025internlm}, R1-Reward \citep{zhang2025r1}, MM-RLHF-Reward \citep{zhang2025mm}, and Unified-Reward-Think \citep{wang2025unifiedthink}; and (3) our own ablation baselines Proxy-GRM-SFT (cold-start only) and Proxy-GRM-RL (RL without proxy reward).

\begin{table*}[t]
\centering
\caption{Combined Results on VL-RewardBench, Multimodal Reward Bench, and MM-RLHF-Reward Bench. Best results are marked with a \best{green background} and second best with a \secondbest{yellow background}.}
\label{tab:combined_results}
\resizebox{\textwidth}{!}{%
\begin{tabular}{l l c S[table-format=2.2] S[table-format=2.2] S[table-format=2.2] S[table-format=2.2] S[table-format=2.2]}
\toprule
\multirow{2}{*}{\textbf{Model}} & \multirow{2}{*}{\textbf{Proxy Agent}} & {\textbf{Data}} & \multicolumn{2}{c}{\textbf{VL-RewardBench}} & {\textbf{Multimodal}} & \multicolumn{2}{c}{\textbf{MM-RLHF-Reward Bench}} \\
\cmidrule(lr){4-5} \cmidrule(lr){6-6} \cmidrule(lr){7-8}
 & & {\textbf{Size}} & {\textbf{Overall\textsubscript{Acc}}} & {\textbf{Macro\textsubscript{Acc}}} & {\textbf{Reward Bench}} & {\textbf{Acc}} & {\textbf{Acc+}} \\
\midrule

\rowcolor{mygray} 
\multicolumn{8}{c}{\textbf{Proprietary Models}} \\
GPT-4o-(2024-08-06) & None & {-} & 65.80 & 62.40 & 70.80 & 58.23 & 26.01 \\
Claude-3.5-Sonnet-(2024-06-22) & None & {-} & 55.30 & 53.60 & 71.50 & 62.94 & 26.11 \\
Claude-3.7-Sonnet & None & {-} & 66.31 & 66.53 & 71.90 & \secondbest{82.35} & \best{65.22} \\
\midrule

\rowcolor{mygray} 
\multicolumn{8}{c}{\textbf{Open-source Models}} \\
VITA-1.5 & None & {-} & 16.48 & 16.53 & 53.60 & 20.58 & 2.78 \\
SliME & None & {-} & 19.04 & 17.64 & 42.00 & 17.10 & 1.76 \\
NVLM-D-72B & None & {-} & 40.10 & 44.10 & {--} & 34.70 & 6.52 \\
Llama-3.2-90B & None & {-} & 56.20 & 53.90 & {--} & 35.29 & 10.86 \\
Qwen2-VL-72B & None & {-} & 39.50 & 43.00 & 70.90 & 48.23 & 13.04 \\
IXC-2.5-Reward & None & {>200k} & 65.80 & 70.00 & 66.60 & 71.18 & 50.00 \\
R1-Reward & None & {>200k} & 71.92 & 71.44 & 82.20 & 80.59 & 54.35 \\
Unified-Reward-SFT & None & {>200k} & 66.10 & 66.50 & {--} & {--} & {--} \\
Unified-Reward-Think & None & {>200k} & \secondbest{73.80} & 72.30 & {--} & {--} & {--} \\
\midrule

\rowcolor{mygray} 
\multicolumn{8}{c}{\textbf{Our Baselines}} \\
Proxy-GRM-SFT & None & {10k} & 69.53 & 69.18 & 85.18 & 79.41 & 52.17 \\
Proxy-GRM-RL & None & {45k} & 72.17 & 71.18 & \secondbest{85.28} & \best{82.94} & 56.52 \\
\midrule
Proxy-GRM-RL & Proxy-SFT & {50k} & \best{75.22} & \best{73.93} & \best{85.62} & \best{82.94} & 56.52 \\
Proxy-GRM-RL & Proxy-RL & {60k} & 73.38 & \secondbest{72.38} & 84.78 & 81.76 & \secondbest{58.70} \\
\bottomrule
\end{tabular}%
}
\end{table*}

\subsection{Main Results}
We present the main comparison results on three benchmarks, demonstrating that Proxy-GRM achieves state-of-the-art performance with significantly less training data. Combined results are shown in \Cref{tab:combined_results}, and the detailed results for each benchmark are shown in the \Cref{sec:detailed_results}.

\noindent\textbf{Results on VL-RewardBench.} As shown in \Cref{tab:combined_results}, Proxy-GRM-RL with Proxy-SFT as the proxy agent achieves the highest overall accuracy of 75.22\% and macro accuracy of 73.93\%, surpassing all existing methods. Compared to the previous best open-source model, Unified-Reward-Think (73.8\% overall), our method yields a 1.42 percentage point improvement. Against R1-Reward (71.92\%), the margin widens to 3.30 points. Moreover, Proxy-SFT consistently outperforms Proxy-RL as a proxy agent. We investigate this surprising finding in detail in \Cref{sec:analysis_studies}.

\noindent\textbf{Results on Multimodal Reward Bench.} On this benchmark, Proxy-GRM-RL with Proxy-SFT achieves the best overall score of 85.62\%, outperforming R1-Reward (82.2\%) by 3.42 points and surpassing all closed-source models including Claude-3.7-Sonnet (71.9\%) and GPT-4o (70.8\%) by substantial margins. The consistent improvements across categories—including challenging dimensions such as coding and mathematics—suggest that proxy-guided rubric generation learns evaluation strategies that are fundamentally more reliable, rather than overfitting to specific task patterns. 

\noindent\textbf{Results on MM-RLHF-Reward Bench.} On this benchmark, Proxy-GRM-RL with Proxy-SFT achieves the highest Acc of 82.94\%, surpassing R1-Reward (80.59\%) and all closed-source models except Claude-3.7-Sonnet on the Acc+ metric. The progressive improvement from Proxy-GRM-SFT (79.41\%) to Proxy-GRM-RL (82.94\%) clearly demonstrates the complementary benefits of reinforcement learning and proxy-guided rubric verification. 

\noindent\textbf{Data efficiency.} A key advantage of Proxy-GRM is its data efficiency. While R1-Reward and Unified-Reward-RL are trained on over 200k samples, Proxy-GRM uses only ~50k samples, which is roughly 4$\times$ less data. Despite this significant reduction, Proxy-GRM consistently outperforms these baselines, suggesting that proxy-guided rubric verification provides a more efficient learning signal than simply scaling training data.

\begin{figure*}[t]
  \centering
  \includegraphics[width=\textwidth]{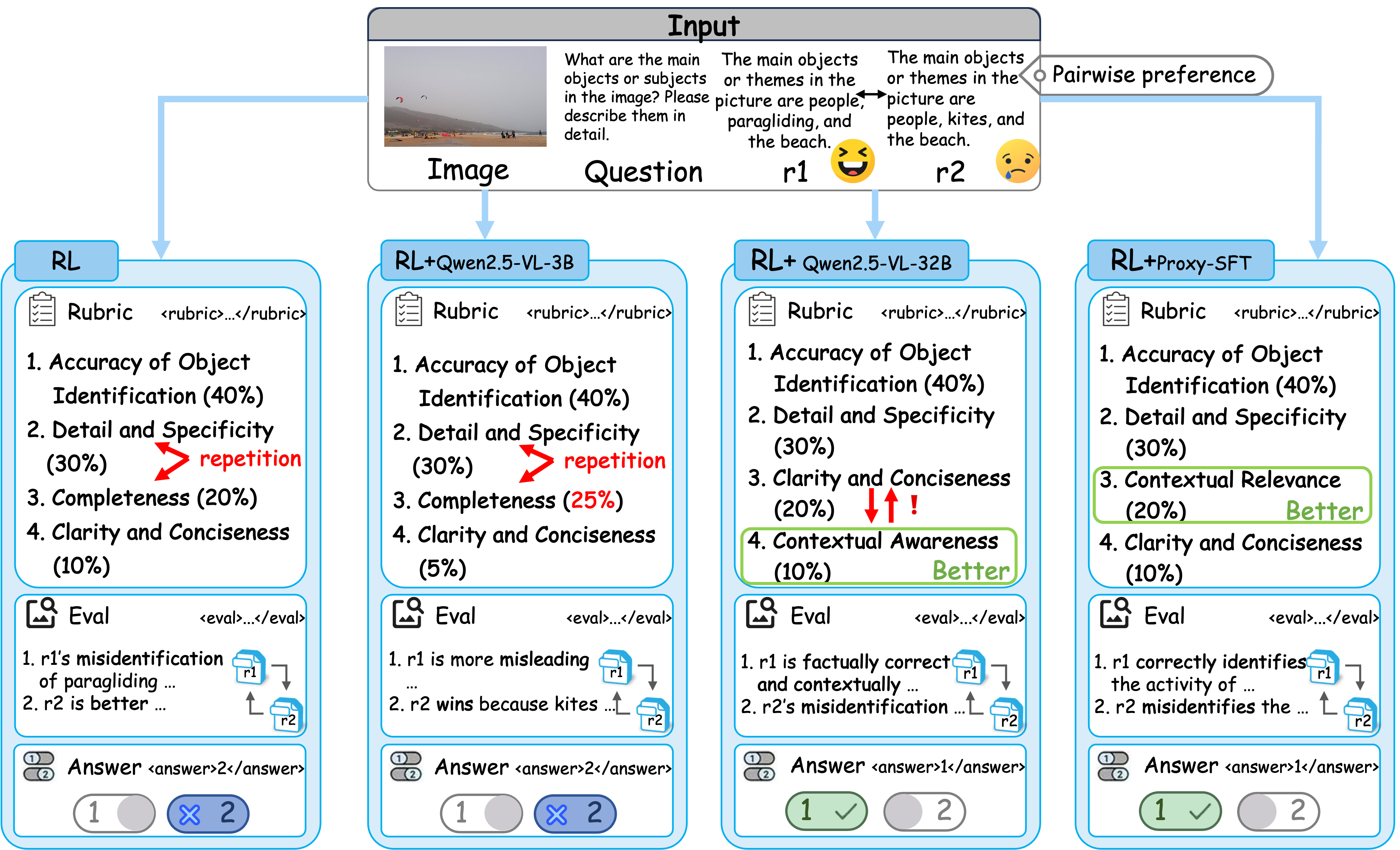} 
  \caption{\textbf{Qualitative comparison of rubrics generated under different proxy agent configurations.}  Given the same multimodal input with a pairwise preference (r1 is preferred), we compare four variants: \textbf{RL} (no proxy) produces rubrics with redundant criteria and an incorrect verdict; \textbf{RL+Qwen2.5-VL-3B} fails to correct repetition and distorts weight allocation, also yielding an incorrect answer; \textbf{RL+Qwen2.5-VL-32B} introduces a more informative criterion (\emph{Contextual Awareness}) and arrives at the correct verdict; \textbf{RL+Proxy-SFT} generates the most discriminative rubric with \emph{Contextual Relevance}, balanced weights, and a correct prediction. Stronger proxy agents guide the policy toward more specific and non-redundant rubrics.}
  \label{fig:case}
\end{figure*}

\begin{table}[t]
  \centering
  \caption{Proxy Agent Selection Results using \textbf{Proxy-GRM-RL} as the base model. Best results are marked with a \best{green background}.}
  \label{tab:ProxyAgentSelection}
  \scalebox{0.9}{%
  \begin{tabular}{l S[table-format=2.2] S[table-format=2.2] S[table-format=2.2] S[table-format=2.2]}
  \toprule
  \textbf{Proxy Agents} & {\textbf{Avg Score}} & 
  {\makecell[c]{\textbf{VL-Reward}\\\textbf{Bench}}} &
  {\makecell[c]{\textbf{Multimodal}\\\textbf{Reward Bench}}} &
  {\makecell[c]{\textbf{MM-RLHF-}\\\textbf{Reward Bench}}} \\
  \midrule
  
  Proxy-SFT & \best{81.26} & \best{75.22} & \best{85.62} & \best{82.94} \\
  Proxy-RL & 79.97 & 73.38 & 84.78 & 81.76 \\
  Unified-Reward-SFT & 81.18 & 74.98 & \best{85.62} & \best{82.94} \\
  Unified-Reward-Think & 79.77 & 73.14 & 85.00 & 81.18 \\
  R1-Reward & 80.59 & 73.54 & 85.28 & \best{82.94} \\
  Qwen2.5-VL-3B-Instruct & 77.87 & 72.25 & 84.88 & 76.48 \\
  Qwen2.5-VL-7B-Instruct & 79.76 & 73.70 & 84.98 & 80.60 \\
  Qwen2.5-VL-32B-Instruct & 80.25 & 74.02 & 84.98 & 81.76 \\
  \bottomrule
  \end{tabular}%
  }
\end{table}

\subsection{Analysis Studies}\label{sec:analysis_studies}
We conduct extensive analysis studies to understand the design choices underlying Proxy-GRM.

\noindent\textbf{Proxy Agent Selection.} A natural question is what makes a good proxy agent. We systematically compare three categories of proxy agents: (1) general-purpose instruction-following models of varying sizes (Qwen2.5-VL-3B/7B/32B-Instruct), (2) existing reward models (R1-Reward, Unified-Reward-SFT, Unified-Reward-Think), and (3) our dedicated rubric evaluation agents (Proxy-SFT, Proxy-RL). Table~\ref{tab:ProxyAgentSelection} summarizes the results.

\textbf{SFT vs. RL proxy agents.} A consistent finding across both our agents and external reward models is that SFT-based models outperform RL-based models as proxy agents. Proxy-SFT outperforms Proxy-RL on all three benchmarks (75.22\% vs. 73.38\% on VL-RewardBench). Similarly, Unified-Reward-SFT outperforms Unified-Reward-Think (74.98\% vs. 73.14\%).

\begin{table}[t]
\centering
\caption{Reward feedback configurations. Each row specifies the composite reward value for the four possible combinations of accuracy reward ($r_{acc}$) and proxy reward ($r_{proxy}$) outcomes. The rightmost column indicates whether $r_{proxy}$ is ultimately zeroed out in the final reward.}
\label{tab:rewardconfig}
\scalebox{0.9}{%
\begin{tabular}{l cccc c}
\toprule
\multirow{2}{*}{\textbf{Config}} & \multicolumn{2}{c}{$r_{acc}=1$} & \multicolumn{2}{c}{$r_{acc} \neq 1$} & \multirow{2}{*}{{\begin{tabular}[c]{@{}c@{}}\textbf{Zero-out}\\$r_{proxy}$\end{tabular}}} \\
\cmidrule(lr){2-3} \cmidrule(lr){4-5}
& $r_{proxy}=1$ & $r_{proxy} \neq 1$ & $r_{proxy}=1$ & $r_{proxy} \neq 1$ & \\
\midrule
\texttt{fb}$_1$ & $1.5$ & $0.5$ & $-1.0$ & $-1.0$ & \redx \\
\texttt{fb}$_2$ & $1.5$ & $0.5$ & $-1.5$ & $-1.0$ & \redx \\
\texttt{fb}$_3$ & $1.5$ & $0.5$ & $-1.5$ & $-2.0$ & \redx \\
\texttt{fb}$_4$ & $1.0$ & $0.5$ & $-1.0$ & $-1.0$ & \redx \\
\texttt{fb}$_5$ & $1.5$ & $0.5$ & $-1.0$ & $-1.0$ & \greencheck \\
\texttt{fb}$_6$ & $1.5$ & $0.5$ & $-1.5$ & $-1.0$ & \greencheck \\
\texttt{fb}$_7$ & $1.5$ & $0.5$ & $-1.5$ & $-2.0$ & \greencheck \\
\bottomrule
\end{tabular}%
}
\end{table}

\begin{figure}[t]
  \centering
  \includegraphics[width=\textwidth]{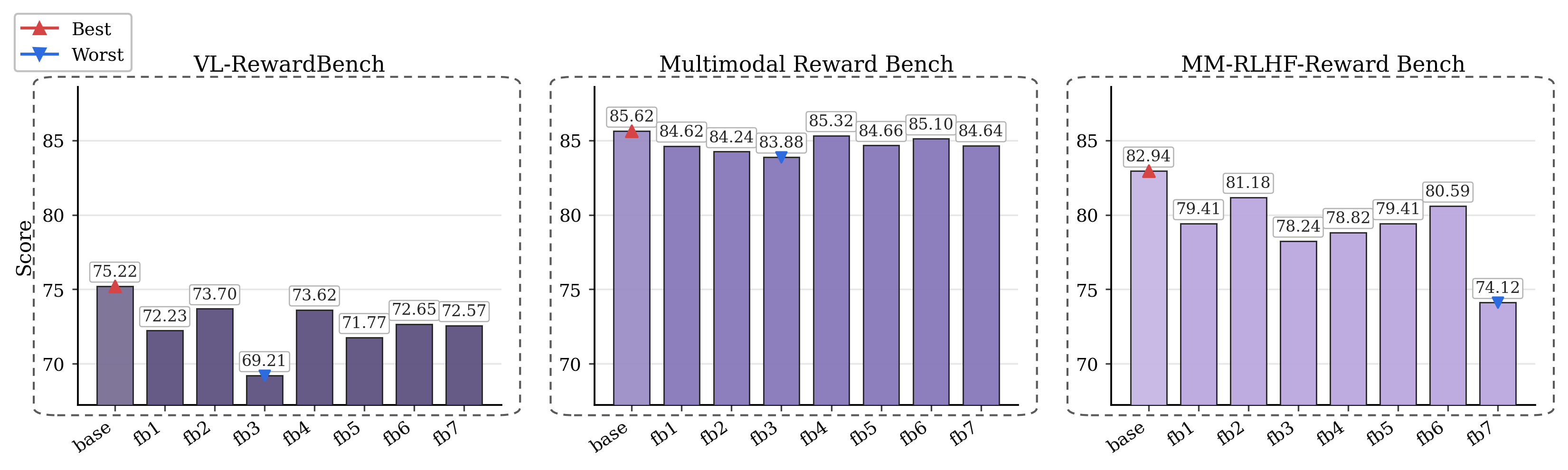}
  \caption{\textbf{Comparison of average scores across different Proxy Reward Feedback on three benchmarks.} Each subplot reports the mean score for each variant, with numeric annotations above bars. Triangular markers indicate the best and worst performing variants within each benchmark. Overall, the base model achieves the highest scores on VL-RewardBench and Instructability, while performance on MM-RLHF-RewardBench remains consistently high with smaller variation across variants.}
  \label{fig:rewardfb}
\end{figure}

We attribute this to a fundamental tension in outcome-supervised RL: models trained with only final-answer rewards may develop evaluation processes that are internally inconsistent, producing correct verdicts through flawed reasoning, or generating plausible-looking evaluations that do not faithfully reflect their actual decision process. When such models are used as rubric evaluators, their inconsistent evaluation processes lead to noisy and unreliable rubric quality signals. In contrast, SFT models are trained to faithfully reproduce the teacher's evaluation process, resulting in more reliable rubric-following behavior. Notably, R1-Reward, an RL-trained model that incorporates some process-level supervision during training, achieves higher proxy performance (73.54\%) than Proxy-RL and Unified-Reward-Think, providing further evidence that process-level consistency is key to effective rubric evaluation.

\textbf{Effect of model scale.} Among general-purpose instruction models, larger models yield better proxy performance: Qwen2.5-VL-32B-Instruct (74.02\% on VL-RewardBench) outperforms Qwen2.5-VL-7B-Instruct (73.70\%), which in turn outperforms Qwen2.5-VL-3B-Instruct (72.25\%). However, the 3B model actually degrades performance below the no-proxy baseline on MM-RLHF-Reward Bench (76.48\% vs. 80.59\%), indicating that insufficiently capable proxy agents introduce harmful noise into the training signal. As illustrated in \Cref{fig:case}, the 3B model fails to distinguish between high-quality and low-quality rubrics during early RL training, providing essentially random feedback that misleads the policy model about what constitutes a good rubric.

\textbf{Dedicated vs. general-purpose agents.} Proxy-SFT achieves the best overall performance despite being based on a 7B model, outperforming even the much larger Qwen2.5-VL-32B-Instruct on VL-RewardBench (75.22\% vs. 74.02\%). This demonstrates the value of specialized training for rubric evaluation, and the 5k targeted training samples instill rubric-following capabilities that general instruction tuning does not provide. As shown in \Cref{fig:case}, although the 32B model introduces a new rubric, its understanding of the critical aspects of the rubric is weaker than that of Proxy-SFT.

\noindent\textbf{Reward Configuration Analysis.} Our default configuration uses implicit reward aggregation. We investigate whether explicitly modifying the accuracy reward based on proxy feedback could be more effective than simply adding the two signals. Table~\ref{tab:rewardconfig} defines seven alternative feedback configurations that modify \(r_{acc}\) based on the proxy outcome, with varying degrees of penalty for rubric-inconsistent predictions.

\Cref{fig:rewardfb} presents the results. All explicit modification schemes perform worse than the implicit additive baseline. The most aggressive penalty scheme (\(fb3\) when both accuracy and proxy fail) causes training instability, while moderate schemes (\(fb1, fb4\)) underperform the additive baseline by 1-3 points. Zeroing out \(r_{proxy}\) after modifying \(r_{acc}\) (configurations \(fb5-fb7\)) performs comparably to keeping it, but neither approach matches the simple additive formulation.

We hypothesize that explicit reward modification creates credit assignment confusion: when \(r_{acc}\) is modified based on \(r_{proxy}\), the policy cannot clearly distinguish whether a reward change stems from its final verdict or its rubric quality, hindering efficient learning of both components. The additive formulation preserves clean credit assignment by keeping the two signals independent.

\begin{table}[t]
  \centering
  \caption{Rubric Transferability Results on VL-Reward-Bench, Multimodal Reward Bench, MM-RLHF-Reward Bench.}
  \label{tab:rubricverify}
  \resizebox{\linewidth}{!}{
  \begin{tabular}{llccc}
  \toprule
  \textbf{Evaluator} & \textbf{Rubric Source} &
  \makecell{\textbf{VL-Reward}\\\textbf{Bench}} &
  \makecell{\textbf{Multimodal}\\\textbf{Reward Bench}} &
  \makecell{\textbf{MM-RLHF-}\\\textbf{Reward Bench}} \\
  \midrule
  \multirow{6}{*}{Qwen2.5-VL-7B-Instruct} 
  & Qwen2.5-VL-7B-Instruct & 36.89 & 70.98 & 40.00 \\
  & Qwen2.5-VL-32B-Instruct & 41.06 & 71.08 & 41.18 \\
  & UnifiedReward-SFT & 42.02 & 72.36 & 41.76 \\
  & Proxy-GRM-RL & 52.33 & 79.46 & 71.03 \\
  & Proxy-GRM-RL + UnifiedReward-SFT & 53.87 & 79.82 & 73.79 \\
  & Proxy-GRM-RL + Proxy-SFT & \best{54.13} & \best{80.00} & \best{73.79} \\
  \midrule
  
  \multirow{6}{*}{Qwen2.5-VL-32B-Instruct} 
  & Qwen2.5-VL-7B-Instruct & 40.02 & 62.46 & 40.59 \\
  & Qwen2.5-VL-32B-Instruct & 43.30 & 72.48 & 42.35 \\
  & UnifiedReward-SFT & 44.91 & 73.84 & 43.53 \\
  & Proxy-GRM-RL & 66.37 & 80.02 & 70.83 \\
  & Proxy-GRM-RL + UnifiedReward-SFT & 66.63 & 81.02 & 73.10 \\
  & Proxy-GRM-RL + Proxy-SFT & \best{66.80} &  \best{81.14} & \best{75.17} \\
  \midrule
  
  \multirow{6}{*}{UnifiedReward-SFT} 
  & Qwen2.5-VL-7B-Instruct & 49.00 & 63.08 & 58.82 \\
  & Qwen2.5-VL-32B-Instruct & 51.32 & 65.16 & 60.00 \\
  & UnifiedReward-SFT & 63.56 & 72.88 & 67.65 \\
  & Proxy-GRM-RL & 67.84 & 82.16 & 75.18 \\
  & Proxy-GRM-RL + UnifiedReward-SFT & 68.60 & 82.92 & 77.06 \\
  & Proxy-GRM-RL + Proxy-SFT & \best{69.86} & \best{83.14} & \best{77.65} \\
  \bottomrule
  \end{tabular}
  }
\end{table}

\subsection{Rubric Transferability Analysis}
A central claim of our approach is that proxy-guided training produces rubrics that are not merely self-consistent but genuinely transferable, which means that the rubrics should be useful for external models that did not generate them. We validate this through a rubric transfer experiment.

\textbf{Protocol.} For each sample in the evaluation benchmarks, we extract the rubrics R generated by various rubric sources. These rubrics are then provided to three evaluator models: Qwen2.5-VL-7B-Instruct, Qwen2.5-VL-32B-Instruct, and Unified-Reward-SFT. We measure whether the externally provided rubrics improve their preference accuracy.

\textbf{Results.} Table~\ref{tab:rubricverify} presents the rubric transfer results. Across all three evaluator models and all three benchmarks, rubrics generated by Proxy-SFT consistently provide the best accuracy. For example, when Unified-Reward-SFT uses rubrics from Proxy-GRM-RL with Proxy-SFT, its VL-RewardBench accuracy improves from 66.10\% (with no generated rubrics) to 69.86\%, which is a 3.76 point gain.

Several patterns emerge from the transfer analysis. First, Proxy-SFT rubrics consistently outperform rubrics from all other sources, including Qwen2.5-VL-32B-Instruct (a 4.5$\times$ larger model), confirming that our proxy-guided training genuinely improves rubric quality rather than merely teaching the policy model to exploit specific evaluation shortcuts. Second, the improvements generalize across evaluator models of different architectures and sizes, demonstrating that the rubrics capture universal evaluation principles rather than model-specific patterns. Third, the gains on MM-RLHF-Reward Bench are particularly large, suggesting that transferable rubrics are most valuable for challenging evaluation scenarios where evaluator models would otherwise struggle.

These results validate our core thesis: by optimizing rubrics for external verifiability through proxy agents, we produce evaluation criteria that are genuinely more informative and applicable, rather than merely self-serving rationalizations.

\section{Conclusion}
We have presented Proxy-GRM, a framework for training multimodal generative reward models that produce robust, transferable evaluation rubrics. By introducing proxy agents that verify rubric quality during RL training, we address a fundamental limitation of existing approaches: the lack of any quality signal for the intermediate rubric generation step. Our key findings include: (1) proxy-guided rubric verification consistently improves reward model performance across three benchmarks, achieving state-of-the-art results with 4$\times$ less training data than comparable methods; (2) SFT-based proxy agents outperform RL-based ones for rubric evaluation, highlighting the importance of process fidelity in verification tasks; (3) simple implicit reward aggregation is more effective than explicitly modifying the accuracy reward based on proxy feedback; and (4) rubrics generated by Proxy-GRM transfer effectively to unseen evaluator models, validating that our approach genuinely improves rubric transferability. We believe that the principle of external verifiability, which involves optimizing intermediate reasoning artifacts to be useful for independent agents, represents a promising direction for improving the reliability and interpretability of generative reward models.

\bibliography{conference}
\bibliographystyle{conference}

\appendix
\clearpage
\section{Prompt}
\subsection{Instruction Prompt}\label{sec:ins_prompt}

\begin{figure}[!htbp]
  \centering
  \includegraphics[width=\textwidth]{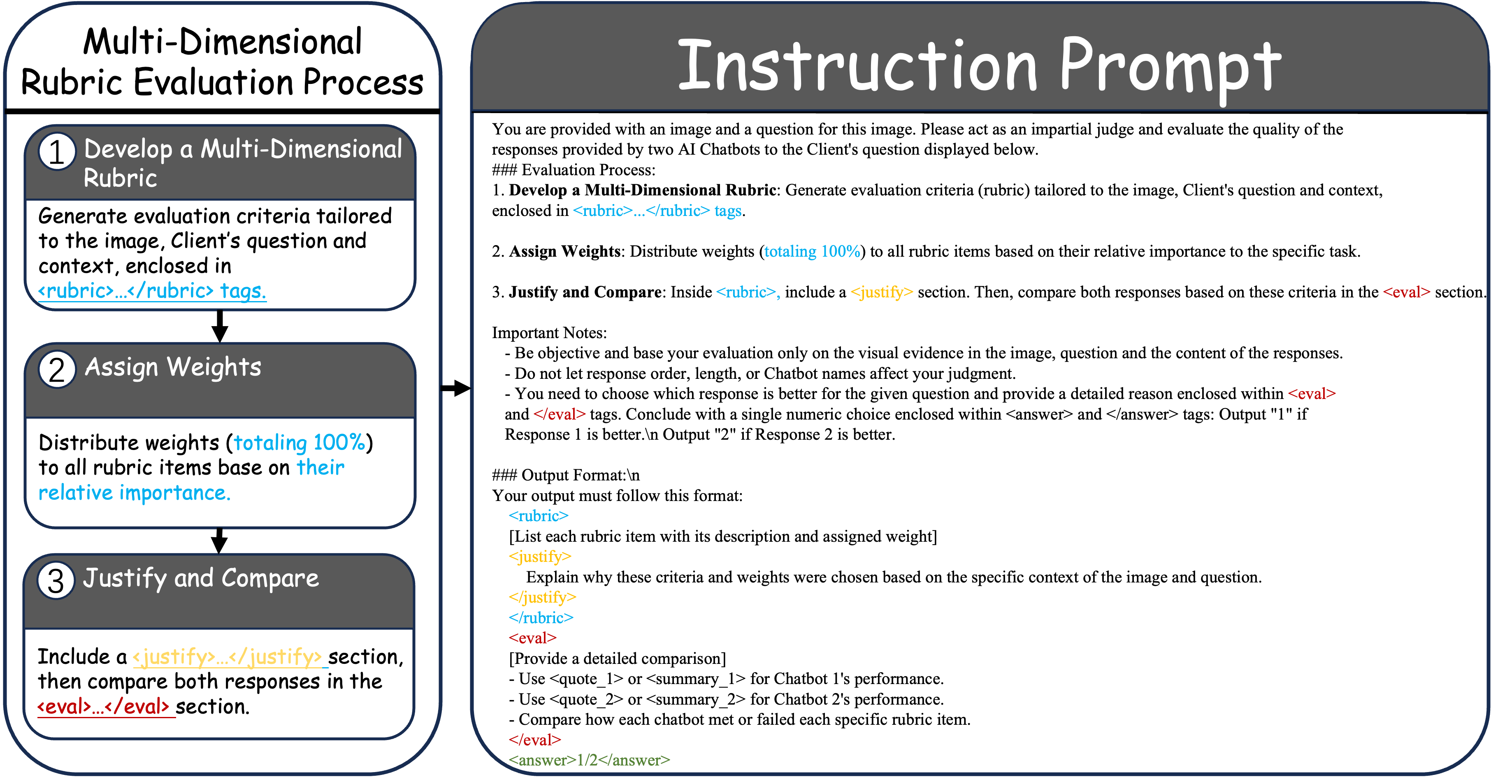}
  \caption{\textbf{System Prompt for Distillation and RL.} }
  \label{fig:instruction_prompt}
\end{figure}

\subsection{Rubric Evaluation Prompt}

\begin{figure}[!b]
  \centering
  \includegraphics[width=0.75\textwidth]{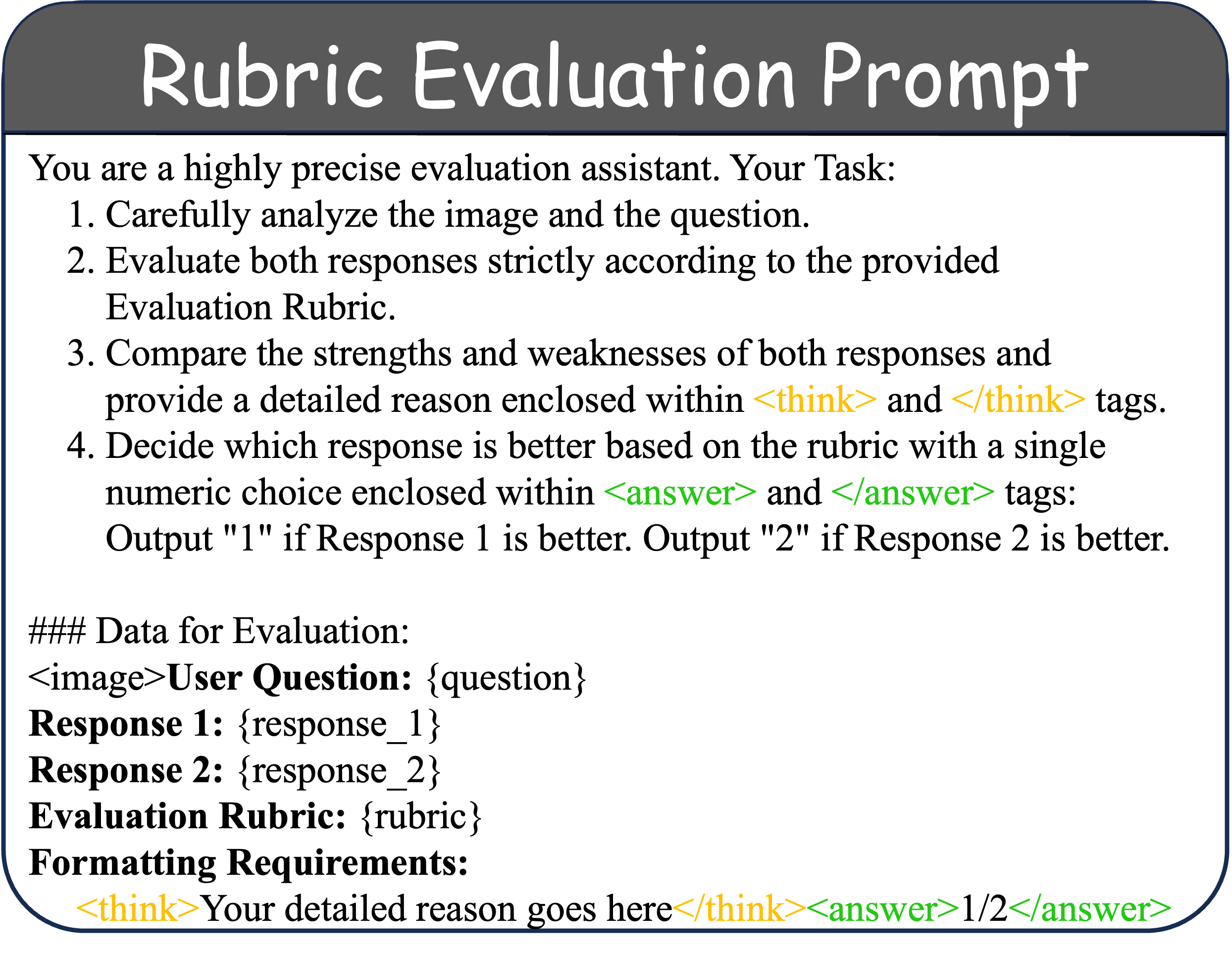}
  \caption{\textbf{System Prompt for Rubric Evaluation.} }
  \label{fig:evaluation_prompt}
\end{figure}

\section{Training Datasets}
\subsection{Data Sources and Curation}
We curate training data from four established multimodal preference datasets: LLaVA-Critic-113k, RLAIF-V, RLHF-V, and MMIF-23k, yielding approximately 60k samples in total after quality filtering. To ensure data diversity and difficulty, we apply three automatic filter criteria:
\begin{itemize}
    \item \textbf{Quality:} We remove samples with degenerate or trivially distinguishable response pairs.
    
    \item \textbf{Difficulty:} We retain samples where the preference distinction is non-trivial, to provide meaningful learning signal.
    
    \item \textbf{Similarity:} We deduplicate samples based on semantic similarity to avoid redundancy. 
\end{itemize}

\subsection{Data Distillation}
For each of the 60k curated samples, we use Qwen3-VL-235B-A22B as the teacher model to generate structured critiques in the $\texttt{<rubric>}\mathcal{R}\texttt{</rubric><eval>}\mathcal{E}\texttt{</eval><answer>}\mathcal{A}\texttt{</answer>}$ format. The distillation prompt in \Cref{sec:ins_prompt} instructs the teacher to produce fine-grained evaluation criteria, apply them systematically, and output a final verdict.

\subsection{Data Allocation}
From the 60k distilled outputs, we select samples where the teacher model produces the correct preference, yielding approximately 25k correct samples. The remaining 35k samples (including those where the teacher was incorrect) serve as hard negative data for RL training. The 25k correct samples are allocated in \Cref{tab:data_allocation}.


\begin{table}[!htbp]
\caption{Data allocation for Proxy model training and Proxy-GRM Training.}
\label{tab:data_allocation}
\centering
\begin{tabular}{|m{4cm}|m{1.8cm}|m{8cm}|}
\hline
\textbf{Split} & \textbf{Size} & \textbf{Purpose} \\
\hline
Proxy-SFT training & 5k & Supervised fine-tuning of the proxy agent \\
\hline
Proxy-RL training & 10k & Reinforcement learning of the proxy agent \\
\hline
Proxy-GRM-SFT & 10k & Supervised fine-tuning initialization of the policy model \\
\hline
\end{tabular}
\end{table}

Combined with the 35k hard samples used for Proxy-GRM-RL training, the total RL training pool comprises approximately 45k samples. Including the 5k proxy training samples, the full Proxy-GRM-RL pipeline consumes 50k samples in total, which is roughly $4\times$ less than comparable state-of-the-art methods (e.g., R1-Reward and Unified-Reward-Think, each trained on $>$200k samples).

\section{Training Details}
\subsection{Proxy-GRM-SFT}
The policy model's cold-start SFT stage is implemented using ms-swift. Key hyperparameters:
\begin{table}[!htbp]
\label{tab:sft_params}
\centering
\begin{tabular}{|l|c|}
\hline
\textbf{Hyperparameter} & \textbf{Value}  \\
\hline
Training framework & ms-swift \\
\hline
Epochs & 1  \\
\hline
Learning rate & \(1 \times 10^{-5}\)  \\
\hline
LR schedule & Cosine decay  \\
\hline
Training samples & 10k  \\
\hline
\end{tabular}
\end{table}

\subsection{Proxy-GRM-RL}
The reinforcement learning stage is implemented using verl with GRPO. Key hyperparameters:

\begin{table}[!htbp]
\label{tab:rl_params}
\centering
\begin{tabular}{|l|c|}
\hline
\textbf{Hyperparameter} & \textbf{Value}  \\
\hline
Training framework & verl \\
\hline
Epochs & 1  \\
\hline
Learning rate & \(5 \times 10^{-6}\)  \\
\hline
Rollout group size \(G\) & 7  \\
\hline
Training batch size & 256  \\
\hline
Mini-batch size & 128 \\
\hline
Proxy Agent & Frozen \\
\hline
\end{tabular}
\end{table}

Both the proxy agent training stages (Proxy-SFT and Proxy-RL) follow analogous configurations, with Proxy-SFT using the same SFT setup as the policy cold-start and Proxy-RL using the same GRPO setup as Proxy-GRM-RL.

\section{Detailed Experimental Results}\label{sec:detailed_results}

The following subsections provide benchmark-level breakdowns of all models evaluated in this work, supplementing the condensed \Cref{tab:combined_results} in the main paper.

\subsection{Results on VL-RewardBench}
VL-RewardBench contains 1,247 preference pairs spanning three sub-categories: General multimodal understanding, Hallucination detection, and visual Reasoning. We report both Overall Accuracy and Macro Accuracy in \Cref{tab:vl_rewardbench}.

Proxy-GRM-RL with Proxy-SFT achieves the best overall accuracy of $75.22\%$, outperforming Unified-Reward-Think by 1.42 points and R1-Reward by 3.30 points. Notably, our model shows particularly strong gains on the Hallucination sub-category ($93.08\%$ vs. $85.71\%$ for R1-Reward), suggesting that proxy-guided rubric generation encourages the policy to produce evaluation criteria that are especially sensitive to factual grounding. The Reasoning sub-category remains competitive but shows less pronounced gains, consistent with the observation in the main paper that rubric transferability provides the most value in scenarios where fine-grained evaluation criteria matter.

\subsection{Results on Multimodal Reward Bench}
Multimodal Reward Bench evaluates reward models across seven dimensions: Correctness, Preference, Knowledge, Math, Coding, Safety, and VQA, with 5,000 total samples. We report the overall accuracy and per-category breakdowns in \Cref{tab:mm_rb}.

On this benchmark, both Proxy-GRM-SFT and Proxy-GRM-RL already exceed all prior methods by a substantial margin, with our strongest configuration (Proxy-GRM-RL + Proxy-SFT / Unified-Reward-SFT) achieving 85.6\% overall—a 3.4-point improvement over R1-Reward (82.2\%) and a 13–15 point improvement over all closed-source models. Notably, the performance differences among proxy agent variants are smaller on this benchmark than on VL-RewardBench, suggesting that the structured nature of Multimodal Reward Bench categories is largely addressable by the base Proxy-GRM-RL model, with proxy guidance providing incremental refinement.

\subsection{Results on MM-RLHF-Reward Bench}
MM-RLHF-Reward Bench is a challenging benchmark covering Multiple-Choice (MCQ), Long-form, Short-form, Safety, and Video evaluation tasks. We report the standard accuracy (Acc) and the stricter Acc+ metric, which requires correctness on the harder subset of questions. The detailed results are shown in \Cref{tab:mm_rlhf_rb}.

Our method achieves the best Acc of 82.94\% across three configurations (Proxy-SFT, Unified-Reward-SFT, and R1-Reward as proxy). On the stricter Acc+ metric, Proxy-GRM-RL with R1-Reward achieves 63.04\%, matching Claude-3.7-Sonnet and substantially outperforming R1-Reward (54.35\%). The progressive improvement from Proxy-GRM-SFT (79.41\%) \(\rightarrow\) Proxy-GRM-RL (80.59\%) \(\rightarrow\) Proxy-GRM-RL + Proxy-SFT (82.94\%) demonstrates the complementary contributions of RL training and proxy-guided rubric verification. A noteworthy observation is that the Qwen2.5-VL-3B proxy agent causes a notable performance degradation (76.47\%), falling below even the SFT-only baseline, consistent with our finding in Section 4.3 that insufficiently capable proxy agents introduce harmful training noise.

\begin{table*}[!htbp]
\centering
\caption{Results on VL-RewardBench. Best results are marked with a \best{green background} and second best with a \secondbest{yellow background}.}
\label{tab:vl_rewardbench}
\resizebox{\linewidth}{!}{
\begin{tabular}{l l c c c c c}
\midrule
\textbf{Model} & \textbf{Proxy Agent} & $\textbf{Overall}_\text{Acc}$ & $\textbf{Macro}_\text{Acc}$ & \textbf{General} & \textbf{Hallucination} & \textbf{Reasoning} \\
\midrule
\rowcolor{mygray}
\multicolumn{7}{c}{\textbf{Proprietary Models}} \\
\midrule
GPT-4o (2024-08-06) & None & 65.80 & 62.40 & 49.10 & 67.60 & \best{70.50} \\
Claude-3.5-Sonnet (2024-06-22) & None & 55.30 & 53.60 & 43.40 & 55.00 & 62.30 \\
Claude-3.7-Sonnet & None & 66.31 & 66.53 & 68.08 & 70.70 & 60.81 \\
Gemini-1.5-Flash (2024-09-24) & None & 57.60 & 55.30 & 47.80 & 59.60 & 58.40 \\
Gemini-1.5-Pro (2024-09-24) & None & 67.20 & 62.50 & 50.80 & 72.50 & 64.20 \\
\midrule
\rowcolor{mygray}
\multicolumn{7}{c}{\textbf{Open-source Reward Models}} \\
\midrule
VITA-1.5 & None & 16.48 & 16.53 & 18.55 & 8.93 & 22.11 \\
SliME & None & 19.04 & 17.64 & 7.23 & 27.09 & 18.60 \\
LLaVA-OneVision-7B-ov & None & 29.60 & 36.50 & 32.20 & 20.10 & 57.10 \\
Molmo-7B & None & 37.50 & 39.70 & 31.10 & 31.80 & 56.20 \\
InternVL2-8B & None & 44.50 & 45.20 & 35.60 & 41.10 & 59.00 \\
LLaVA-Critic-8B & None & 41.20 & 44.00 & 54.60 & 38.30 & 59.10 \\
Llama-3.2-11B & None & 42.90 & 42.80 & 33.30 & 38.40 & 56.60 \\
Pixtral-12B & None & 35.80 & 40.40 & 35.60 & 25.90 & 59.90 \\
Molmo-72B & None & 44.10 & 43.70 & 33.90 & 42.30 & 54.90 \\
Qwen2-VL-72B & None & 39.50 & 43.00 & 38.10 & 32.80 & 58.00 \\
NVLM-D-72B & None & 40.10 & 44.10 & 38.90 & 31.60 & 62.00 \\
MM-RLHF-Reward & None & 50.15 & 51.01 & 45.04 & 50.45 & 57.55 \\
Llama-3.2-90B & None & 56.20 & 53.90 & 42.60 & 57.30 & 61.70 \\
IXC-2.5-Reward & None & 65.80 & 70.00 & 84.70 & 62.50 & 62.90 \\
R1-Reward & None & 71.92 & 71.44 & 63.84 & 85.71 & 64.78 \\
Unified-Reward-SFT & None & 66.10 & 66.50 & 60.60 & 78.40 & 60.50 \\
Unified-Reward-Think & None & 73.80 & 72.30 & \best{78.10} & 72.70 & \secondbest{66.00} \\
\midrule
\rowcolor{mygray} 
\multicolumn{7}{c}{\textbf{Our Baselines}} \\
\hline
Proxy-GRM-SFT & None & 69.53 & 69.18 & 57.26 & 87.50 & 62.78 \\
Proxy-GRM-RL & None & 72.17 & 71.18 & 61.83 & 92.41 & 59.31 \\
\midrule
\textbf{Proxy-GRM-RL} & \textbf{Proxy-SFT} & \best{75.22} & \secondbest{73.93} & \secondbest{68.46} & \best{93.08} & 60.25 \\
Proxy-GRM-RL & Proxy-RL & 73.38 & 72.38 & 64.52 & 91.74 & 60.88 \\
Proxy-GRM-RL & Unified-Reward-SFT & \secondbest{74.98} & \best{74.32} & 64.94 & 92.41 & 65.62 \\
Proxy-GRM-RL & Unified-Reward-Think & 73.14 & 72.13 & 64.52 & 91.29 & 60.57 \\
Proxy-GRM-RL & R1-Reward & 73.54 & 72.93 & 63.28 & 90.85 & 64.67 \\
Proxy-GRM-RL & Qwen2.5-VL-3B-Instruct & 72.25 & 71.57 & 61.41 & 90.85 & 62.46 \\
Proxy-GRM-RL & Qwen2.5-VL-7B-Instruct & 73.70 & 73.01 & 62.45 & \secondbest{92.86} & 63.72 \\
Proxy-GRM-RL & Qwen2.5-VL-32B-Instruct & 74.02 & 73.25 & 63.49 & \secondbest{92.86} & 63.41 \\
\midrule
\end{tabular}
}
\end{table*}

\begin{table*}[!htbp]
\centering
\caption{Results on Multimodal Reward Bench. Best results are marked with a \best{green background} and second best with a \secondbest{yellow background}.}
\label{tab:mm_rb}
\resizebox{\linewidth}{!}{
\begin{tabular}{l l c c c c c c c c}
\midrule
\textbf{Model} & \textbf{Proxy Agent} & \textbf{Overall} & \textbf{Correctness} & \textbf{Preference} & \textbf{Knowledge} & \textbf{Math} & \textbf{Coding} & \textbf{Safety} & \textbf{VQA} \\
\midrule
\rowcolor{mygray}
\multicolumn{10}{c}{\textbf{Proprietary Models}} \\
\midrule
GPT-4o (2024-08-06) & None & 70.8 & 62.6 & 69.0 & 72.0 & 67.6 & 62.1 & 74.8 & 87.2 \\
Gemini-1.5-Pro (2024-09-24) & None & 71.9 & 63.5 & 67.7 & 66.3 & 68.9 & 55.5 & 94.5 & 87.2 \\
Claude-3.5-Sonnet (2024-06-22) & None & 71.5 & 62.6 & 67.8 & 73.9 & 68.6 & 65.1 & 76.8 & 85.6 \\
Claude-3.7-Sonnet & None & 71.9 & 58.4 & 60.7 & 78.1 & 76.3 & 71.3 & 72.0 & 86.8 \\
\midrule
\rowcolor{mygray}
\multicolumn{10}{c}{\textbf{Open-source Reward Models}} \\
\midrule
SliME & None & 42.0 & 42.3 & 52.2 & 47.5 & 43.5 & 35.3 & 19.1 & 53.8 \\
VITA-1.5 & None & 53.6 & 55.6 & 54.3 & 52.5 & 51.9 & 52.8 & 58.1 & 50.0 \\
Llama-3.2-Vision-Instruct (11B) & None & 51.2 & 57.8 & 65.8 & 55.5 & 50.6 & 51.7 & 20.9 & 55.8 \\
Molmo-D-0924 & None & 52.9 & 56.8 & 59.4 & 54.6 & 50.7 & 53.4 & 34.8 & 60.3 \\
Llama-3.2-Vision-Instruct (90B) & None & 61.2 & 60.0 & 68.4 & 61.2 & 56.3 & 53.1 & 52.0 & 77.1 \\
InternVL-3 & None & 63.6 & 59.6 & 61.6 & 60.5 & 65.1 & 56.6 & 59.3 & 82.3 \\
Qwen2-VL & None & 70.9 & 56.4 & 62.3 & 70.2 & 73.3 & 58.9 & 90.1 & 85.3 \\
MM-RLHF-Reward & None & 67.1 & 61.7 & 67.5 & 54.3 & 58.4 & 57.9 & 92.9 & 76.8 \\
IXC-2.5-Reward & None & 66.6 & 60.7 & 64.2 & 56.8 & 63.0 & 50.5 & 89.9 & 81.1 \\
R1-Reward & None & 82.2 & 77.5 & 74.0 & 74.9 & \secondbest{83.1} & \best{79.6} & \best{99.6} & \best{86.5} \\
\midrule
\rowcolor{mygray}
\multicolumn{10}{c}{\textbf{Our Baselines}} \\
\midrule
Proxy-GRM-SFT & None & 85.2 & 87.2 & 84.7 & 86.0 & 82.0 & \secondbest{78.7} & \secondbest{96.8} & 83.6 \\
Proxy-GRM-RL & None & \secondbest{85.3} & 87.0 & \best{86.6} & 85.6 & 83.0 & 78.6 & 94.5 & 83.8 \\
\midrule
\textbf{Proxy-GRM-RL} & \textbf{Proxy-SFT} & \best{85.6} & 87.2 & 84.9 & \best{86.3} & 82.7 & 78.2 & 96.7 & \secondbest{85.0} \\
Proxy-GRM-RL & Proxy-RL & 84.8 & 86.8 & 84.7 & 85.8 & 81.2 & 75.9 & 95.9 & 84.4 \\
Proxy-GRM-RL & Unified-Reward-SFT & \best{85.6} & \best{87.7} & 85.2 & 86.1 & \best{83.6} & 77.9 & 96.0 & 84.7 \\
Proxy-GRM-RL & Unified-Reward-Think & 85.0 & 86.5 & \secondbest{85.4} & 85.5 & 81.7 & 77.2 & 95.4 & 84.5 \\
Proxy-GRM-RL & R1-Reward & 85.2 & 85.7 & \secondbest{85.4} & \best{86.5} & 82.6 & 77.0 & 96.0 & 84.8 \\
Proxy-GRM-RL & Qwen2.5-VL-3B-Instruct & 84.9 & 85.9 & 84.5 & 85.9 & 81.7 & 77.4 & 95.8 & 84.4 \\
Proxy-GRM-RL & Qwen2.5-VL-7B-Instruct & 84.9 & 86.5 & 84.1 & 85.5 & 82.9 & 76.6 & 96.5 & 84.1 \\
Proxy-GRM-RL & Qwen2.5-VL-32B-Instruct & 84.9 & \secondbest{87.3} & 85.0 & 84.9 & 83.0 & 77.0 & 96.0 & 83.6 \\
\midrule
\end{tabular}
}
\end{table*}

\begin{table*}[!htbp]
\centering
\caption{Results on MM-RLHF-Reward Bench. Best results are marked with a \best{green background} and second best with a \secondbest{yellow background}.}
\label{tab:mm_rlhf_rb}
\resizebox{\linewidth}{!}{
\begin{tabular}{l l c c c c c c c}
\midrule
\textbf{Model} & \textbf{Proxy Agent} & \textbf{Acc} & \textbf{Acc+} & \textbf{MCQ} & \textbf{Long} & \textbf{Short} & \textbf{Safety} & \textbf{Video} \\
\midrule
\rowcolor{mygray}
\multicolumn{9}{c}{\textbf{Proprietary Models}} \\
\midrule
Gemini-2.0-Flash-Exp & None & 44.71 & 13.04 & 33.33 & 45.94 & 67.64 & 43.75 & 32.00 \\
GPT-4o (2024-08-06) & None & 58.23 & 26.01 & 64.28 & 78.37 & 44.11 & 56.25 & 40.00 \\
Claude-3.5-Sonnet (2024-06-22) & None & 62.94 & 26.11 & 64.28 & 67.56 & 55.88 & 65.62 & 60.00 \\
Claude-3.7-Sonnet & None & \secondbest{82.35} & \best{65.22} & 66.67 & 91.89 & \best{91.18} & \best{87.50} & 76.00 \\
\hline
\rowcolor{mygray}
\multicolumn{9}{c}{\textbf{Open-source Reward Models}} \\
\midrule
SliME & None & 17.10 & 1.76 & 23.81 & 10.81 & 14.71 & 12.50 & 7.52 \\
VITA-1.5 & None & 20.58 & 2.78 & 24.97 & 21.62 & 11.76 & 18.75 & 12.40 \\
InternVL-3 & None & 37.65 & 6.52 & 35.71 & 56.76 & 23.53 & 37.50 & 32.00 \\
NVLM-D-72B & None & 34.70 & 6.52 & 42.85 & 32.43 & 8.82 & 50.00 & 40.00 \\
Llama-3.2-90B & None & 35.29 & 10.86 & 19.04 & 35.13 & 38.23 & 50.00 & 40.00 \\
Qwen2-VL-72B & None & 48.23 & 13.04 & 45.23 & 62.16 & 47.05 & 46.88 & 36.00 \\
IXC-2.5-Reward & None & 71.18 & 50.00 & 52.38 & 91.89 & 67.65 & 62.50 & \best{88.00} \\
MM-RLHF-Reward & None & 82.00 & 63.00 & \best{83.00} & 97.00 & 74.00 & 69.00 & \best{88.00} \\
R1-Reward & None & 80.59 & 54.35 & \secondbest{80.95} & 89.19 & \secondbest{82.35} & 75.00 & 72.00 \\
\midrule
\rowcolor{mygray}
\multicolumn{9}{c}{\textbf{Our Baselines}} \\
\midrule
Proxy-GRM-SFT & None & 79.41 & 52.17 & 59.52 & 94.59 & \secondbest{82.35} & 78.12 & \best{88.00} \\
Proxy-GRM-RL & None & 80.59 & 56.52 & 73.81 & 94.59 & 79.41 & \secondbest{84.38} & 68.00 \\
\midrule
\textbf{Proxy-GRM-RL} & \textbf{Proxy-SFT} & \best{82.94} & 56.52 & 76.19 & \secondbest{97.30} & 79.41 & 81.25 & 80.00 \\
Proxy-GRM-RL & Proxy-RL & 81.76 & 58.70 & 78.57 & 91.89 & 76.47 & 81.25 & 80.00 \\
Proxy-GRM-RL & Unified-Reward-SFT & \best{82.94} & 56.52 & \secondbest{80.95} & 94.59 & 76.47 & 78.12 & \secondbest{84.00} \\
Proxy-GRM-RL & Unified-Reward-Think & 81.18 & 58.70 & 73.81 & 91.89 & 76.47 & \secondbest{84.38} & 80.00 \\
Proxy-GRM-RL & R1-Reward & \best{82.94} & \secondbest{63.04} & 69.05 & \best{100.00} & \secondbest{82.35} & 81.25 & \secondbest{84.00} \\
Proxy-GRM-RL & Qwen2.5-VL-3B-Instruct & 76.47 & 47.83 & 73.81 & 91.89 & 64.71 & 75.00 & 76.00 \\
Proxy-GRM-RL & Qwen2.5-VL-7B-Instruct & 80.59 & 52.17 & 71.43 & \secondbest{97.30} & 76.47 & 78.12 & 80.00 \\
Proxy-GRM-RL & Qwen2.5-VL-32B-Instruct & 81.76 & \secondbest{63.04} & 73.81 & \best{100.00} & 73.53 & 75.00 & \best{88.00} \\
\midrule
\end{tabular}
}
\end{table*}

\section{Ablation Studies}
\subsection{Multi-Agent Proxy Configurations}
\textbf{Motivation:} Given that several individual proxy agents yield strong performance, a natural question is whether combining multiple agents as an ensemble can further improve results. To investigate, we experiment with all pairwise and triple combinations of the four best-performing individual agents: Proxy-SFT (PSFT), Unified-Reward-SFT (USFT), Qwen2.5-VL-7B-Instruct (Q7B), and Qwen2.5-VL-32B-Instruct (Q32B). When using multiple proxy agents, the composite proxy reward is computed as the unweighted average of individual proxy rewards:
\begin{equation}
    r_{\text{proxy}} = \frac{1}{K} \sum_{k=1}^{K} r_{\text{proxy}}^{(k)}
\end{equation}

where $K$ is the number of proxy agents in the ensemble.

\begin{figure*}[!htbp]
  \centering
  \includegraphics[width=\textwidth]{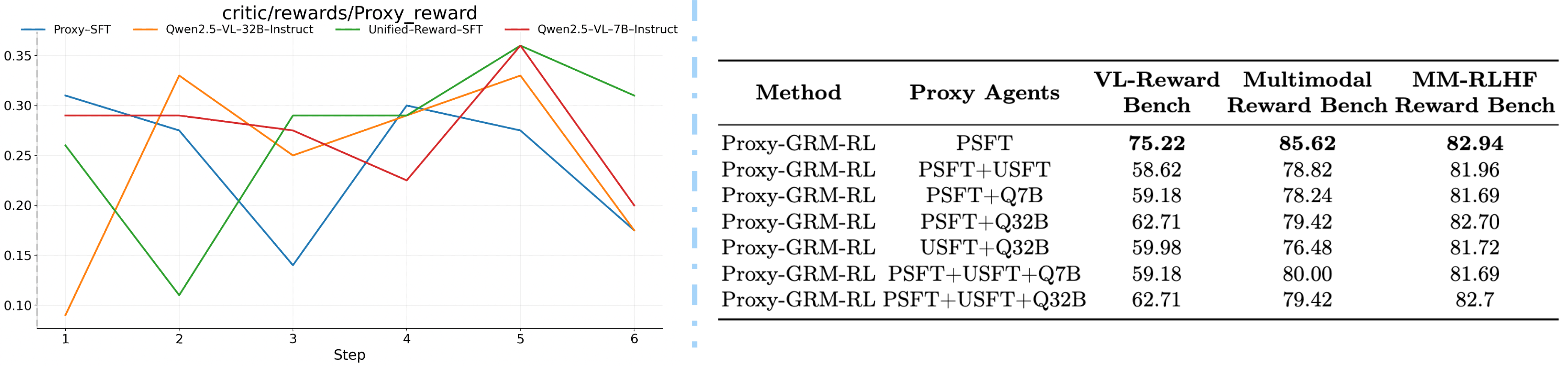}
  \caption{\textbf{Left.} With Proxy-SFT, Unified-Reward-SFT, Qwen2.5-VL-7B-Instruct, or Qwen2.5-VL-32B-Instruct as the sole proxy model, early \(r_{proxy}\) optimization shows high variance and occasional oscillations across updates. This suggests rubric-quality feedback is proxy-sensitive at the start of RL, and the proxy signal can be noisy or poorly calibrated before the policy produces consistently interpretable rubrics. \textbf{Right.} We evaluate Proxy-GRM-RL with different proxy-model combinations across three benchmarks. Contrary to the expectation that ensembling improves verification, performance consistently degrades when combining multiple proxy reward models. This implies heterogeneous proxies may be misaligned; aggregating them can introduce conflicting gradients and hinder learning, making a single well-matched proxy preferable.}
  \label{fig:multi_agents}
\end{figure*}

\noindent\textbf{Results:} The right panel of \Cref{fig:multi_agents} presents the multi-agent results, which are strikingly negative: all multi-agent configurations perform substantially worse than the best single agent. The best combination (PSFT + Q32B) achieves only 62.71\% on VL-RewardBench, compared to 75.22\% for PSFT alone—a dramatic 12.51-point degradation. The trend is consistent across all three benchmarks.

\noindent\textbf{Analysis:} We analyze this failure through the proxy reward trajectories during early RL training, shown in the left panel of Figure E1. Due to differing evaluation capabilities and biases, the agents provide conflicting signals: when PSFT assigns a high reward to a particular rubric, Q7B may assign a low reward, and vice versa. This conflict creates a noisy, ambiguous reward landscape that prevents the policy from learning coherent rubric generation strategies. The averaged reward signal washes out the informative gradient from any individual agent, reducing the proxy reward to near-random noise that hinders rather than helps training.

This finding has important implications beyond our specific setting: it suggests that naively combining multiple reward signals—even from individually strong evaluators—can be catastrophic when the evaluators have systematic disagreements about what constitutes quality. A single, well-calibrated proxy agent is preferable to an ensemble of heterogeneous agents whose evaluation tendencies are misaligned.

\subsection{Ablation on Proxy-SFT Training Data Size}
\textbf{Motivation:} The quality of the proxy agent depends critically on the amount and quality of its training data. Since we lack a dedicated benchmark to directly measure proxy judgment quality (the proxy is evaluated only indirectly via its downstream effect on Proxy-GRM-RL), we conduct an ablation over the training data size to identify the optimal amount of distillation data for Proxy-SFT.

\noindent\textbf{Setup:} We train four Proxy-SFT variants using $2.5\text{k}$, $5\text{k}$, $7.5\text{k}$, and $10\text{k}$ distillation samples, respectively, keeping all other hyperparameters fixed. Each variant is used as the proxy agent for Proxy-GRM-RL, and the final reward model performance on all three benchmarks is reported.

\noindent\textbf{Results:} \Cref{tab:proxy_train_data_size} summarizes the findings. 

\noindent\textbf{Discussion:} The $5\text{k}$ configuration achieves the best performance across all benchmarks except Acc+, where $2.5\text{k}$ and $5\text{k}$ tie. Interestingly, performance degrades when more data ($7.5\text{k}$ or $10\text{k}$) is used. We hypothesize that this is due to the diminishing marginal quality of training samples: as we include more samples, we necessarily include harder or more ambiguous cases where the rubric-following task is less clearly defined, potentially introducing noise that overfits the proxy agent to spurious patterns. This overfitting manifests as miscalibrated proxy rewards during Proxy-GRM-RL training, ultimately degrading the final policy model.

This ablation highlights a key practical insight: because we have no direct proxy evaluation metric, the optimal data size for proxy training must be determined through downstream evaluation of the full Proxy-GRM-RL pipeline. The $5\text{k}$ sweet spot represents a balance between learning sufficient rubric-following capability and avoiding overfitting to training distribution artifacts.

\begin{table}[!htbp]
\centering
\caption{\textbf{Ablation on Proxy-SFT training data size.} Results reported for Proxy-GRM-RL with each Proxy-SFT variant as the proxy agent. Best results are marked with a \best{green background}.}
\label{tab:proxy_train_data_size}
\resizebox{\linewidth}{!}{
\begin{tabular}{l c c c c c}
\hline
\textbf{Data Size} &
$\textbf{VL-RB}~\textbf{Overall}_\text{Acc}$ &
$\textbf{VL-RB}~\textbf{Macro}_\text{Acc}$ &
\textbf{MM-RB Overall} &
\textbf{MM-RLHF-RB Acc} &
\textbf{MM-RLHF-RB Acc+} \\
\hline
2.5k & 74.58 & 73.91 & 85.36 & 81.76 & \best{56.52} \\
\best{5k} & \best{75.22} & \best{73.93} & \best{85.62} & \best{82.94} & \best{56.52} \\
7.5k & 72.81 & 72.15 & 85.48 & 81.18 & 52.17 \\
10k & 73.22 & 72.57 & 85.20 & 80.59 & 52.17 \\
\hline
\end{tabular}
}
\end{table}

\section{More Case Studies}
\begin{figure*}[!htbp]
  \centering
  \includegraphics[width=\textwidth]{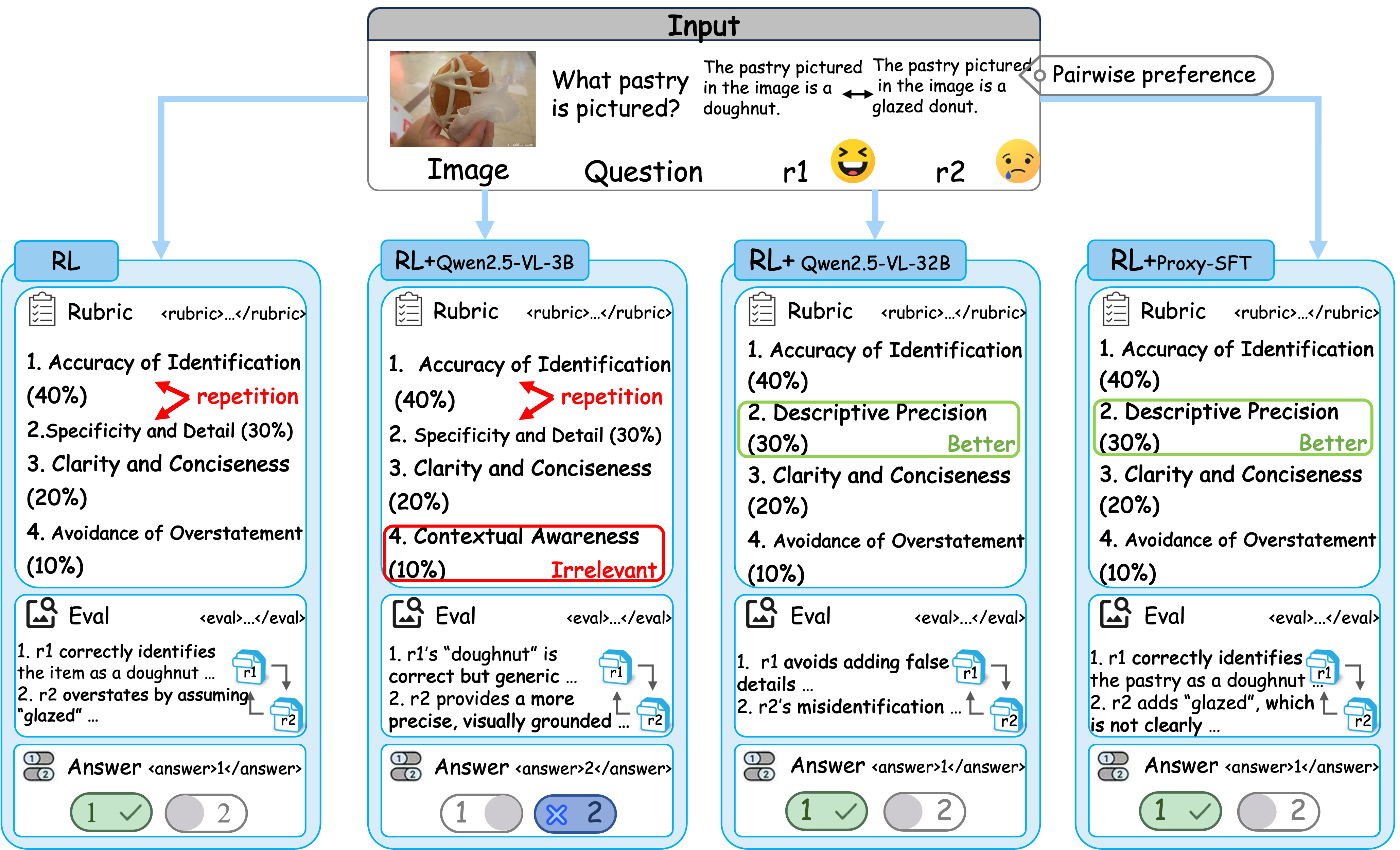} 
  \caption{\textbf{Qualitative comparison of rubrics generated with different proxy configurations.} Given the same multimodal input and a pairwise preference signal (r1 is preferred), we compare four variants: \textbf{RL} (no proxy) produces a rubric with duplicated criteria (e.g., repeated \emph{Accuracy of Identification}), yielding a weakly discriminative evaluation and an incorrect verdict; \textbf{RL+Qwen2.5-VL-3B} still fails to remove repetition, introduces an irrelevant criterion (e.g., \emph{Contextual Awareness}), and misallocates weights, also leading to an incorrect prediction; \textbf{RL+Qwen2.5-VL-32B} replaces redundant/irrelevant items with a more task-aligned and informative criterion (\emph{Descriptive Precision}), improving discriminability and producing the correct verdict; \textbf{RL+Proxy-SFT} generates the most discriminative, non-redundant rubric with better-balanced weights and a correct prediction. Overall, stronger proxy agents guide the policy toward more specific, task-relevant, and non-redundant rubrics.}
  \label{fig:case_1}
\end{figure*}

\begin{figure*}[!htbp]
  \centering
  \includegraphics[width=\textwidth]{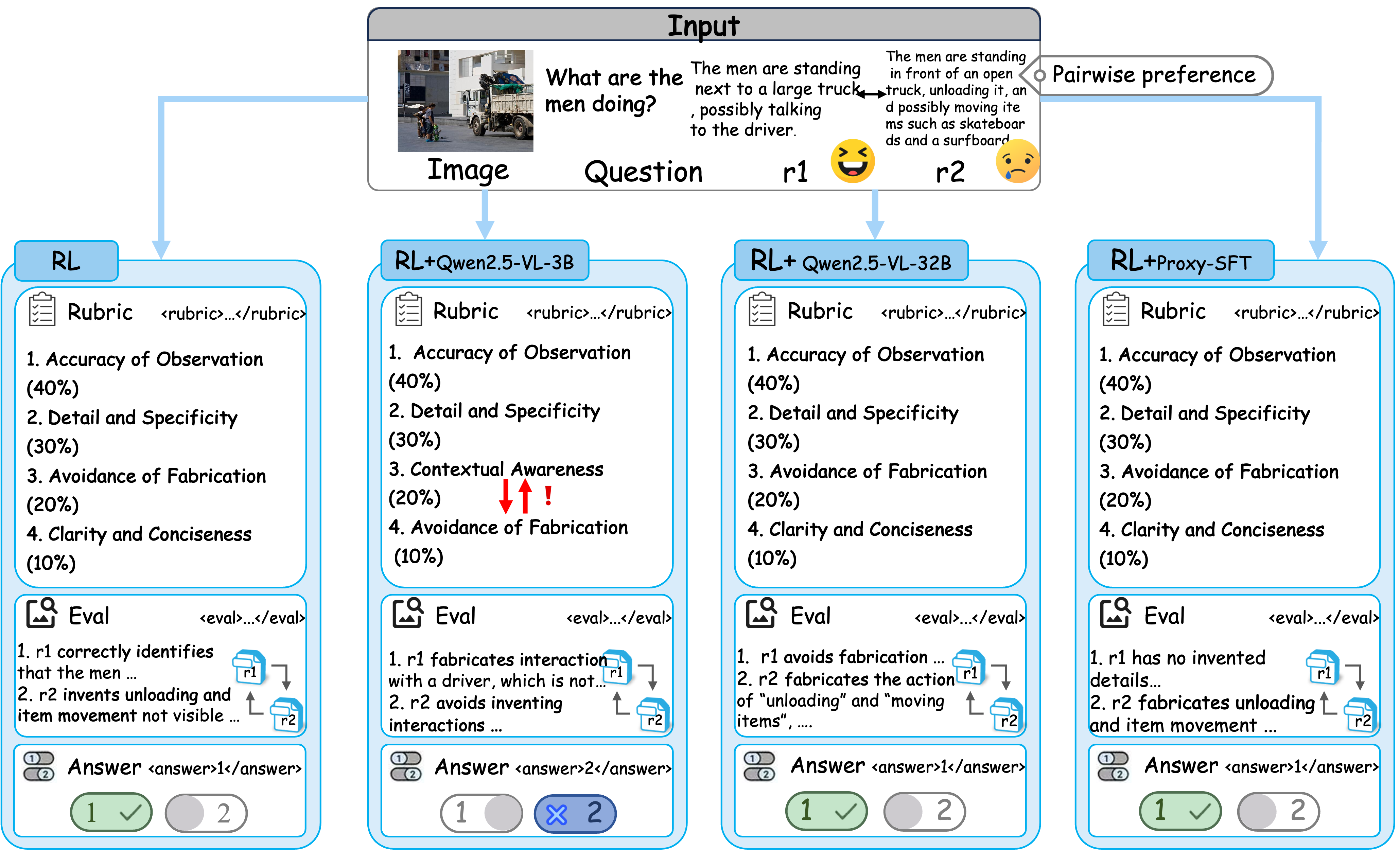} 
  \caption{\textbf{Qualitative comparison of rubrics generated under different proxy agent configurations (action understanding example).} Given the same multimodal input and a pairwise preference signal (r1 is preferred), we compare four variants: \textbf{RL} (no proxy) produces a generic rubric that does not sufficiently penalize hallucinated actions, resulting in an incorrect preference judgment; \textbf{RL+Qwen2.5-VL-3B} over-emphasizes \emph{Contextual Awareness} and shifts weights away from fabrication control, failing to correct the key error and still predicting the wrong winner; \textbf{RL+Qwen2.5-VL-32B} uses a more task-aligned rubric that prioritizes \emph{Avoidance of Fabrication} alongside observation accuracy, correctly favoring r1 over r2’s invented “unloading/moving items” details; \textbf{RL+Proxy-SFT} yields the most robust and balanced rubric, explicitly discouraging invented actions and consistently producing the correct verdict. Overall, stronger proxy agents guide the policy toward more hallucination-sensitive and discriminative rubrics.}
  \label{fig:case_2}
\end{figure*}

\begin{figure*}[!htbp]
  \centering
  \includegraphics[width=\textwidth]{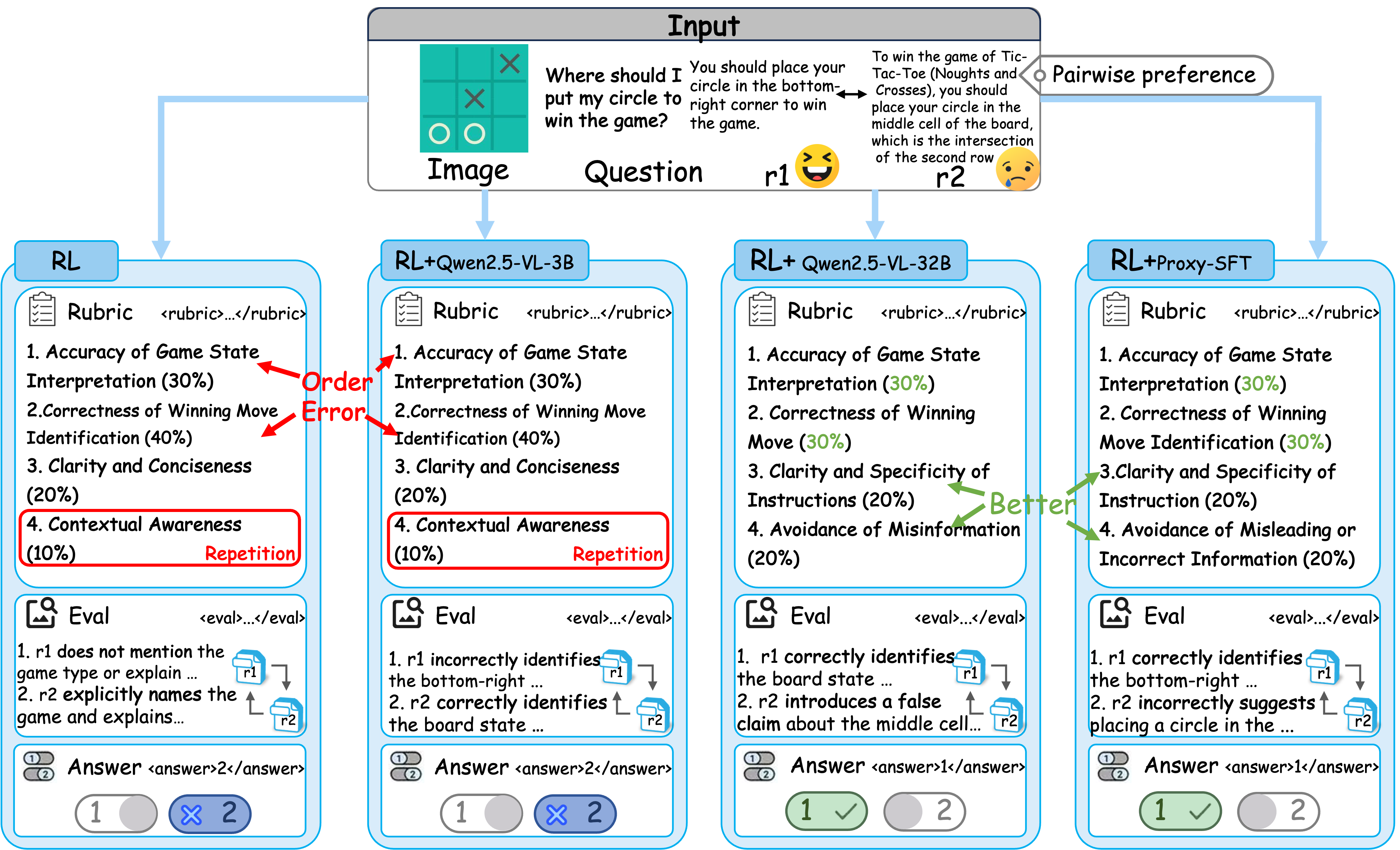} 
  \caption{\textbf{Qualitative comparison of rubrics generated under different proxy agent configurations (game-state reasoning example).} Given the same multimodal input and a pairwise preference signal (r1 is preferred), we compare four variants: \textbf{RL} (no proxy) generates a rubric with an ordering/structure issue and a redundant criterion (repeated \emph{Contextual Awareness}), which weakens the evaluation and leads to an incorrect preference decision; \textbf{RL+Qwen2.5-VL-3B} inherits the same redundancy and fails to emphasize game-state grounding, again producing a wrong verdict; \textbf{RL+Qwen2.5-VL-32B} improves task alignment by focusing on \emph{Accuracy of Game State Interpretation} and \emph{Avoidance of Misinformation}, making the rubric more discriminative and correctly preferring r1 over r2’s incorrect “middle cell” suggestion; \textbf{RL+Proxy-SFT} yields the most balanced and informative rubric, refining instruction quality and explicitly penalizing misleading moves, and thus outputs the correct prediction. Overall, stronger proxy agents steer the policy toward more game-aware, non-redundant rubrics that better detect incorrect strategic advice.}
  \label{fig:case_3}
\end{figure*}

Figure~\ref{fig:case_1}, Figure~\ref{fig:case_2} and Figure~\ref{fig:case_3} present qualitative case studies illustrating how proxy agent strength affects the quality of learned rubrics and the resulting pairwise preference judgments across three representative multimodal tasks (object/attribute description, action understanding, and game-state reasoning). Across all cases, \textbf{RL without a proxy} tends to produce generic or poorly structured rubrics, often containing duplicated or weakly discriminative criteria and misallocated weights, which can yield incorrect winners. Adding a \textbf{small proxy} (Qwen2.5-VL-3B) provides limited benefit and may introduce irrelevant criteria or shift emphasis away from the key failure mode (e.g., hallucination control), frequently failing to fix the prediction. In contrast, a \textbf{stronger proxy} (Qwen2.5-VL-32B) consistently improves task alignment by replacing redundant or irrelevant items with more informative, grounded criteria (e.g., descriptive precision, fabrication avoidance, game-state interpretation), leading to more discriminative rubrics and correct preferences. Finally, \textbf{RL+Proxy-SFT} produces the most robust rubrics overall—non-redundant, better balanced in weighting, and explicitly targeted at task-specific error patterns—resulting in the most reliable preference judgments.

\end{document}